\documentclass[11pt]{article}

\usepackage[final]{acl}

\usepackage{times}
\usepackage{latexsym}

\usepackage[T1]{fontenc}

\usepackage[utf8]{inputenc}

\usepackage{microtype}

\usepackage{inconsolata}

\usepackage{graphicx}

\usepackage{pifont} 
\usepackage{booktabs}

\usepackage{makecell} 
\usepackage{latexsym}
\usepackage{times}
\usepackage{latexsym}
\usepackage{amssymb}
\usepackage{multirow}
\usepackage{threeparttable}
\usepackage{longtable}  
\usepackage{colortbl}
\usepackage[most]{tcolorbox}
\usepackage{subfig}
\usepackage{tablefootnote}
\usepackage{CJKutf8}
\usepackage{url}
\usepackage{footmisc}
\usepackage{diagbox}
\usepackage{xcolor}
\usepackage{enumitem}
\usepackage{hyperref}
\setlist[itemize]{leftmargin=0.8cm} 
\newcommand{\red}{\color{red}}

\definecolor{SectionBg}{gray}{0.95} 
\definecolor{ReasoningBg}{HTML}{E8F0FE} 
\definecolor{OurMethodBg}{HTML}{FFF4E5} 

\title{ReasonTabQA: A Comprehensive Benchmark for Table Question Answering from Real World Industrial Scenarios}

\author{
 \textbf{Changzai Pan\textsuperscript{1,*}},
 \textbf{Jie Zhang\textsuperscript{1,*}},
 \textbf{Kaiwen Wei\textsuperscript{2,*}},
 \textbf{Chenshuo Pan\textsuperscript{1,*}},
 \textbf{Yu Zhao\textsuperscript{1,*}},
\\
 \textbf{Jingwang Huang\textsuperscript{1,*}},
 \textbf{Jian Yang\textsuperscript{3}},
 \textbf{Zhenhe Wu\textsuperscript{1}},
 \textbf{Haoyang Zeng\textsuperscript{2}},
 \textbf{Xiaoyan Gu \textsuperscript{1}},
\\
 \textbf{Weichao Sun\textsuperscript{1}},
 \textbf{Yanbo Zhai\textsuperscript{1}},
 \textbf{Yujie Mao\textsuperscript{1}},
 \textbf{Zhuoru Jiang\textsuperscript{1}},
 \textbf{Jiang Zhong\textsuperscript{2}},
\\
 \textbf{Shuangyong Song\textsuperscript{1}},
 \textbf{Yongxiang Li\textsuperscript{1}},
 \textbf{Zhongjiang He\textsuperscript{1,$\dagger$}} 
\\
 \textsuperscript{1} Institute of Artificial Intelligence (TeleAI), China Telecom,
 \\
 \textsuperscript{2} Chongqing University,
 \textsuperscript{3} Beihang University
}

\begin{document}
\maketitle
\let\thefootnote\relax\footnotetext{* These authors contributed equally to this work.}

\begin{abstract}
Recent advancements in Large Language Models (LLMs) have significantly catalyzed table-based question answering (TableQA). However, existing TableQA benchmarks often overlook the intricacies of industrial scenarios, which are characterized by multi-table structures, nested headers, and massive scales. These environments demand robust table reasoning through deep structured inference, presenting a significant challenge that remains inadequately addressed by current methodologies. To bridge this gap, we present \textbf{ReasonTabQA}, a large-scale bilingual benchmark encompassing 1,932 tables across 30 industry domains such as energy and automotive. 
ReasonTabQA provides high-quality annotations for both final answers and explicit reasoning chains, supporting both thinking and no-thinking paradigms. Furthermore, we introduce \textbf{TabCodeRL}, a reinforcement learning method that leverages table-aware verifiable rewards to guide the generation of logical reasoning paths. Extensive experiments on ReasonTabQA and 4 TableQA datasets demonstrate that while TabCodeRL yields substantial performance gains on open-source LLMs, the persistent performance gap on ReasonTabQA underscores the inherent complexity of real-world industrial TableQA. 
\end{abstract}

\section{Introduction}

\begin{figure}[t]
    \centering 
    \includegraphics[width=0.97\linewidth]{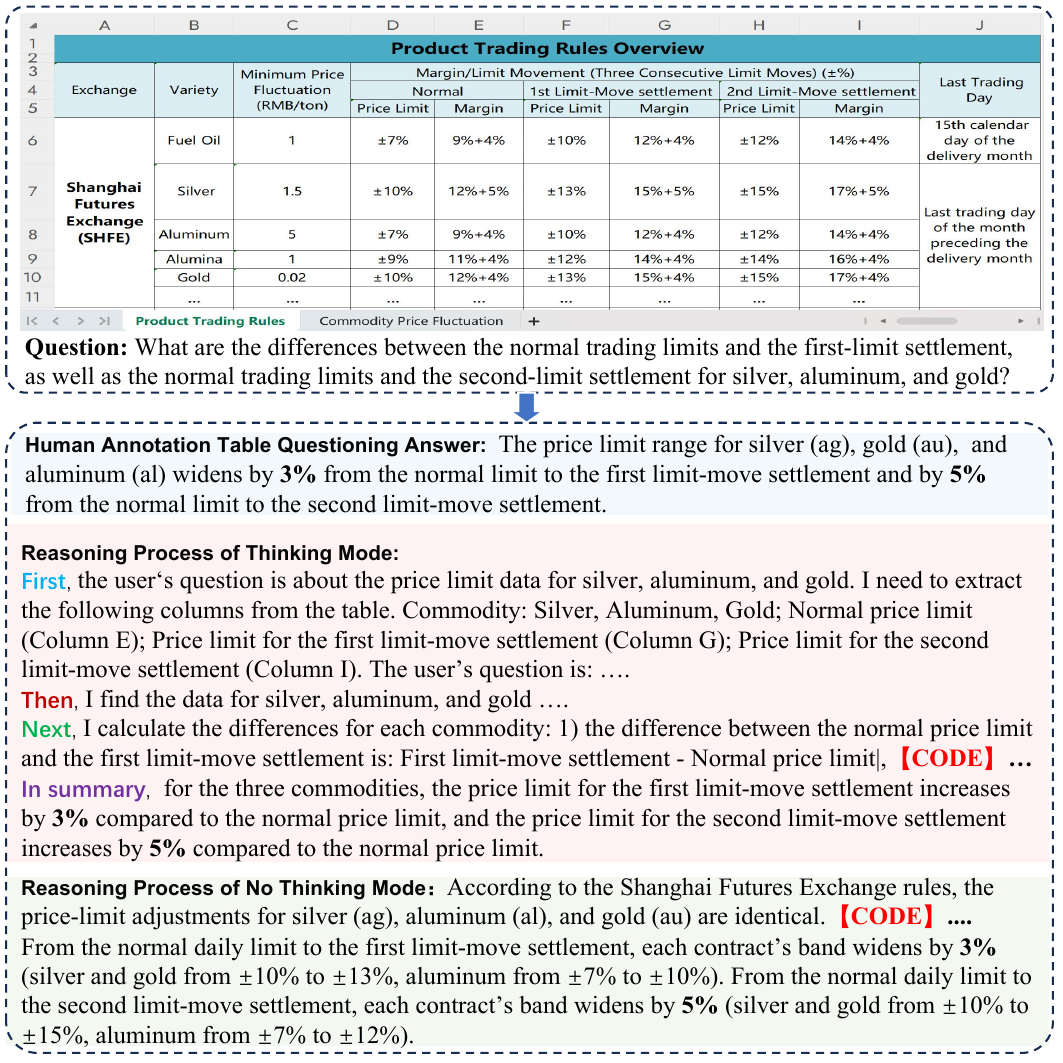} 
    \caption{The ReasonTabQA dataset consists of industrial level tables, annotated questions, annotated gold-standard answers, and annotated reasoning processes across different reasoning modes (thinking and no-thinking). The generated code is omitted.}
    \label{Fig:explation}
\end{figure}


\begin{table*}[!ht]
\centering
\resizebox{0.78\textwidth}{!}{
\begin{tabular}{c c c c c c c c}
    \toprule 
    TableQA Benchmark & \makecell[c]{Multiple\\Tables} & \makecell[c]{Complex\\Structure Tables} & \makecell[c]{Extremely\\Large-Scale Tables} & \makecell[c]{Reasoning Process\\Annotation} & \makecell[c]{Language\\Category} & \makecell[c]{Tables\\Number} & \makecell[c]{Indstury\\Domains}\\
    \midrule 
    TAT-QA \citep{zhu-etal-2021-tat} & $\times$ & $\times$ & $\times$ & $\times$ & en & 20000 & 1\\
    AIT-QA \citep{katsis2022ait} & $\times$ & $\checkmark$ & $\times$ & $\times$ & en & 116 & 1\\
    HiTab \cite{HiTab} & $\times$ & $\checkmark$ & $\times$ & $\times$ & en & 3597 & 29\\
    TableBench \citep{wu2024tablebench} & $\times$ & $\times$ & $\times$ & $\times$ & en & 586 & 18\\
    DataBench \citep{grijalba2024question} & $\times$ & $\times$ & $\checkmark$ & $\times$ & en & 165 & 8\\
    MimoTable \citep{li2024mimotable} & $\checkmark$ & $\checkmark$ & $\times$ & $\times$ & zh, en & 428 & 7\\
    RealHiTBench \citep{wu2025realhitbench} & $\checkmark$ & $\checkmark$ & $\times$ & $\times$ & en & 708 & 24\\
    SciTableQA \citep{ajayi2025scitableqa} & $\checkmark$ & $\checkmark$ & $\times$ & $\times$ & en & 320 & 5 \\
    GRI-QA \citep{contalbo2025gri} & $\checkmark$ & $\checkmark$ & $\times$ &  $\times$ & en & 204 & 7 \\
    MMTU \citep{xing2025mmtu} & $\checkmark$ & $\checkmark$ & $\times$ & $\times$ & en & 61,763 & - \\ 
    T2R-bench \citep{zhang2025t2r} & $\checkmark$ & $\checkmark$ & $\checkmark$ & $\times$ & zh, en & 457 & 19\\
    ReasonTabQA (ours) & $\checkmark$ & $\checkmark$ & $\checkmark$ &  $\checkmark$ & zh, en & 1101(zh), 831(en) & 30\\
    \bottomrule 
\end{tabular}}
\caption{
Comparison between ReasonTabQA and existing representative TableQA benchmarks. Given the scarcity of industrial data in public datasets, we restrict our comparison to benchmarks that incorporate industrial tables.
}
\label{Tab:table_type}
\end{table*}

The rapid evolution of large language models (LLMs) has significantly propelled progress in table reasoning \cite{lu2024large,zhang2024surveytablereasoninglarge,sui2024tablemeetsllmlarge}, spanning diverse tasks such as table-to-text generation \cite{parikh2020totto}, fact verification \cite{chen2019tabfact}, and particularly table question answering (TableQA) \cite{WikiTableQuestion,zhu2021tat,jin2022survey,InfiAgent-DABench,zhang2025rotenhancingtablereasoning}. As a pivotal component of data-driven systems including business intelligence (BI) and enterprise resource planning (ERP), TableQA demands that models perform sophisticated logical inference over structured data to yield precise answers \cite{2020Table,su2024tablegpt2largemultimodalmodel,shi-etal-2024-ehragent}. 

Despite its potential, current TableQA development is primarily constrained by two critical limitations. (1) \textit{from a benchmarking view, existing TableQA datasets fail to capture the complexity of real scenarios, especially in domain coverage and table structural intricacy}. They predominantly focus on open-domain Wikipedia tables or narrow vertical sectors, which fails to represent the diversity of industrial scenarios such as energy, transportation, and trade \cite{WikiTableQuestion, chen2021finqa, nan2022fetaqa, HiTab}. Additionally, these benchmarks often lack structural complexity because they rarely encompass the nested headers, multi-sheet configurations, and extremely large-scale tables pervasive in corporate datasets \cite{li2024mimotable, ma2024spreadsheetbenchchallengingrealworld}. 
(2) \textit{from the method view, current techniques encounter great bottlenecks in industrial TableQA}. 
Existing supervised and program-based methods frequently exhibit grounding and execution errors, such as incorrect table selection, Pandas code generation failures, and unreliable multi-table reasoning, exacerbated by the scarcity of high-quality supervision. Crucially, the absence of table-specific, verifiable reasoning trajectories introduces evaluation blind spots and hinders optimization for real-world complexity \cite{su2024tablegpt2largemultimodalmodel, wang2024chainoftableevolvingtablesreasoning}.




To address these issues, we introduce \textbf{ReasonTabQA}, a comprehensive benchmark specifically engineered for practical industrial scenarios. As summarized in Table~\ref{Tab:table_type}, it comprises 1,101 Chinese and 831 English tables across 30 secondary industrial categories. ReasonTabQA explicitly incorporates 4 complex structural types, including multi-table with multi-sheet configurations, complex and nested headers, extremely large-scale tables, underpinned by a rigorous difficulty taxonomy for both structures and queries. Crucially, as illustrated in Figure~\ref{Fig:explation}, we provide fine-grained annotations of the complete reasoning process in both "thinking" and "no-thinking" modes. These annotations include executable Python code to enhance both transparency and algorithmic robustness.

Furthermore, inspired by Reinforcement Learning with Verifiable Rewards (RLVR) in mathematical and code reasoning \cite{shao2024deepseekmathpushinglimitsmathematical, yu2025dapoopensourcellmreinforcement, deepcoder2025}, we propose \textbf{TabCodeRL}. 
It introduces two table-specific verifiable rewards: a path-selection reward and a code-similarity reward, to provide explicit guidance for generating correct table reasoning paths.
A comprehensive evaluation on ReasonTabQA reveals that even the state-of-the-art model Gemini-3-Pro-Preview achieves an overall score of only 67.58\%. Such results underscore the formidable challenges inherent in industrial-scale table reasoning. Despite these difficulties, the application of TabCodeRL to Qwen3-8B-Instruct \cite{qwen3} yields performance that surpasses all 19 open-source models, validating its effectiveness.
In summary, the contributions of this paper are as follows:


(1) We present ReasonTabQA, a large-scale, bilingual benchmark featuring 1,932 tables across 30 industrial domains. It covers 4 complex structural types and provides fine-grained reasoning path annotations in  thinking and no-thinking modes for model optimization.

(2) We conduct a comprehensive study of 29 models on ReasonTabQA. The results show that even the state-of-the-art Gemini-3-Pro-Preview achieves only 67.58\% overall performance, underscoring the substantial gap between current LLM capabilities and industrial TableQA requirements.


(3) We introduce TabCodeRL, a reinforcement learning framework that leverages table-specific verifiable rewards to enhance industrial tabular reasoning. Experimental results show that it consistently outperforms existing open-source models across multiple TableQA benchmarks.

\begin{figure*}[t!]
    \centering 
    \includegraphics[width=1.0\textwidth]{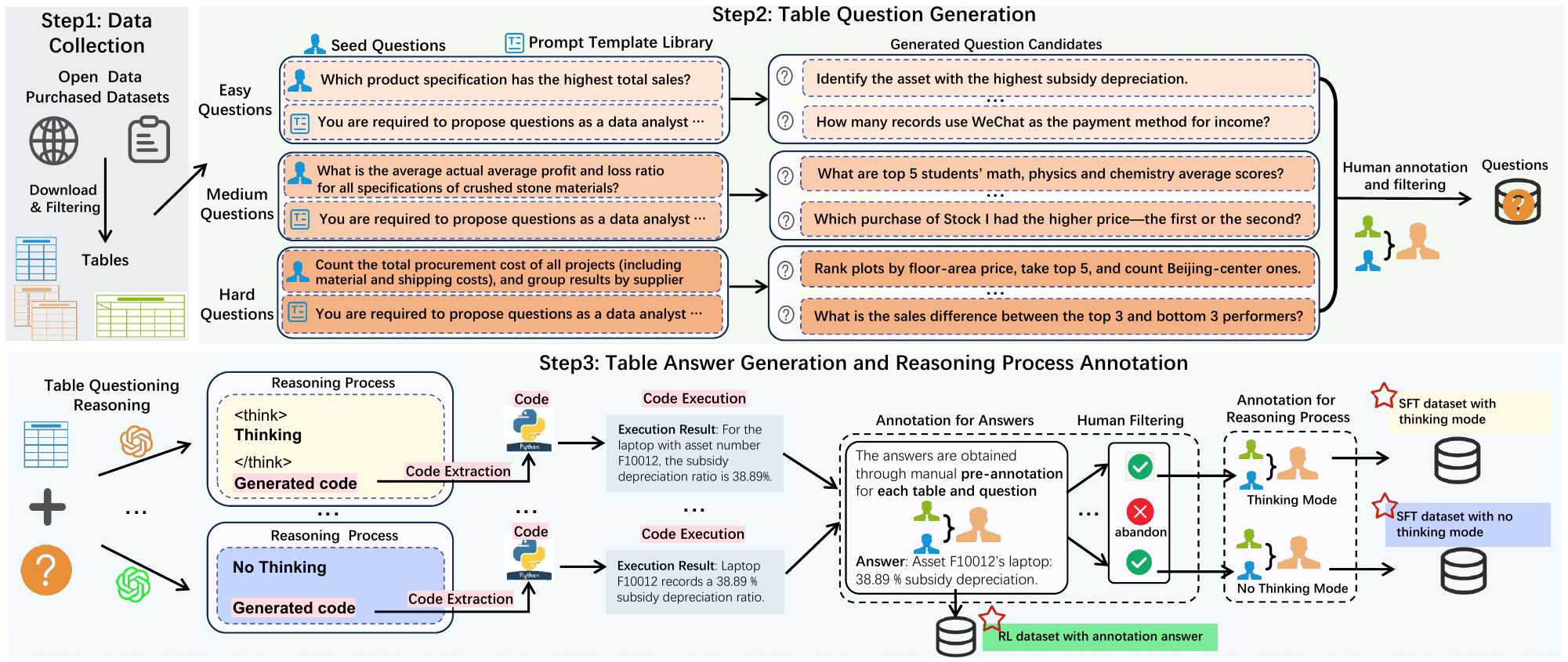} 
    \caption{An overview of the construction pipeline for ReasonTabQA.}
    \label{Fig:bench_pipeline}
\end{figure*}

\section{Related Work}
\paragraph{TableQA Benchmarks.}  
TableQA has evolved from foundational datasets like WTQ \cite{WikiTableQuestion} and TabFact \cite{chen2019tabfact}, which primarily utilize Wikipedia-sourced tables, to more specialized domains such as aviation (AIT-QA \cite{katsis2022ait}), finance (TAT-QA \cite{zhu2021tat}, FinQA \cite{chen2021finqa}) and others \cite{contalbo2025gri, ajayi2025scitableqa}. Recent efforts have begun addressing structural complexity: TableBench \cite{wu2024tablebench} and MMTU \cite{xing2025mmtu} introduces real-world challenges, while MiMoTable \cite{li2024mimotable} and RealHitBench \cite{wu2025realhitbench} explores complex spreadsheet structures. However, existing benchmarks often lack the domain diversity and extreme scale characteristic of industrial environments. Furthermore, most lack manually annotated, verifiable reasoning traces, which are crucial for evaluating and optimizing models in complex, multi-sheet, or large-scale corporate scenarios. We propose ReasonTabQA to fill this void by providing a large-scale, bilingual benchmark with fine-grained reasoning paths across 30 industrial categories.

\paragraph{LLM-based Table Reasoning.}  
Current LLM-based approaches to TableQA generally fall into three categories: (1) prompting-based methods like CoT and Chain-of-Table \cite{wei2023chainofthoughtpromptingelicitsreasoning, wang2024chainoftableevolvingtablesreasoning}; (2) direct fine-tuning on tabular data \cite{zha2023tablegpt, zhang2025tablellmenablingtabulardata}; and (3) program-aided generation where LLMs produce executable code \cite{gao2023palprogramaidedlanguagemodels, wang2024executablecodeactionselicit}. Despite their success, these methods struggle with industrial tables due to structural hallucinations, context window constraints for large-scale data, and a heavy reliance on the inherent coding proficiency of LLMs. To mitigate these limitations, Reinforcement Learning with Verifiable Rewards (RLVR) has emerged as a potent paradigm for enhancing reasoning in math \cite{shao2024deepseekmathpushinglimitsmathematical, wen2025lightr1curriculumsftdpo} and code \cite{le2022coderlmasteringcodegeneration}. In the tabular domain, recent works like Table-R1 \cite{wu2025tabler1regionbasedreinforcementlearning, yang2025tabler1inferencetimescalingtable, yang2025tablegptr1advancingtabularreasoning} and rule-based RL approaches \cite{lei2025reasoningtableexploringreinforcementlearning} demonstrate that verifiable rewards can surpass SFT baselines. Unlike these methods, the proposed TabCodeRL specifically optimizes the alignment between linguistic reasoning and code execution by a table-path selection reward and a code-similarity reward, ensuring more robust and transparent table reasoning paths.

\section{Construction of ReasonTabQA}
\label{Sec:bench construction}

As shown in Figure~\ref{Fig:bench_pipeline}, the construction of ReasonTabQA follows a rigorous pipeline comprising table acquisition, question synthesis, and multi-modal reasoning annotation.

\subsection{Table Acquisition} We curate tables from two primary sources: (1) Public Repositories, including municipal open data platforms, national statistical bureaus, and industry portals; (2) Industrial Reports, with anonymized real-world data and professional service datasets (Appendix~\ref{Appd:source}). To ensure structural representativeness, we incorporate diverse paradigms: multi-sheet configurations, nested hierarchical headers, and extremely large-scale tables. 
Specifically, we employ a two-stage filtering process. First, tables are categorized into 30 secondary classes across 7 domains to ensure domain coverage. Second, we filter out tables with over 50\% null cells and perform manual de-identification to ensure data quality and privacy. Following a rigorous taxonomy, tables are classified into \textit{Simple}, \textit{Medium}, and \textit{Complex} levels (Appendix~\ref{Appd:question_table_difficulty_classification}). The final corpus consists of 1,101 Chinese and 831 English tables.

\begin{table}[t!]
\centering
   \resizebox{0.8\linewidth}{!}{
    \begin{tabular}{ll}
    \toprule
    \textbf{Property} & \textbf{Value} \\ 
    \midrule
    Number of Tables & 1932 \\ 
    Avg Table Files or Sheets for Multi-Tables & 5.04 \\
    Avg Rows per Table & 138.3 \\
    Avg Cells per Table & 1,359.3 \\
    Avg sheets per Directory & 1.5 \\
    Number of Extremely Large-size Tables & 38 \\
    Number of Questions & 5523 \\ 
    Avg Questions per Table & 2.86\\
    Avg Response Length of SFT datast with Thinking & 9366\\
    Avg Response Length of SFT dataset with No Thinking & 1321\\
    \bottomrule
    \end{tabular}}
   \caption{Key Statistics of ReasonTabQA.}
    \label{Tab:key_stat}
\end{table}

\subsection{Table Question Generation}
We adopt a semi-automatic heuristic method to efficiently generate high-quality questions. The specific steps are shown as follows:

\paragraph{Seed Question and Prompt Construction.}  We employed 20 domain experts in data analysis (qualifications detailed in Appendix~\ref{Appd:annotation}) to curate 5 representative seed questions per category. To guide the subsequent generation, these experts also developed a library of prompt templates tailored to each structural difficulty level, ensuring that the questions remain contextually relevant to the table's complexity and ensuring the diversity of questions (refer to Appendix~\ref{Appd:prompts_library_and_seed_questions_for_question generation}).

\paragraph{Self-Instruct Generation.} We utilize the self-instruct paradigm \cite{wang-etal-2023-self-instruct} to scale the question pool. Specifically, we leverage GPT-4o as the backbone generator. For each target table, 3 templates are randomly sampled from our curated library, each incorporating 3--5 expert seeds as in-context demonstrations. This setup instructs the model to generate 3 diverse questions that are both computationally solvable and semantically varied, thereby expanding the breadth of the benchmark.

\paragraph{Human Annotation and Filtering.} To ensure the benchmark meets industrial-grade standards, each candidate question undergoes a rigorous cross-validation process. Two independent annotators evaluate each question against 3 core criteria: (1) tabular answer ability, ensuring the query is self-contained within the table; (2) uniqueness of answer, to avoid ambiguous ground truths; and (3) semantic clarity, ensuring alignment with natural user intent (detailed criteria in Appendix~\ref{Appd:annotation}). In cases of inter-annotator disagreement, a third senior adjudicator is involved for final arbitration (Appendix~\ref{Appd:question_annotation_procedure}). This procedure ultimately yielded 5,523 high-quality questions.

\subsection{Table Answer Generation} 
We employ a Python code generation method \cite{zhang2023reactableenhancingreacttable} to derive ground truth. We deploy 6 LLMs (including QwQ-32B, Qwen2.5-72B-Instruct, Mistral-123B, Qwen3-32B-Instruct, Kimi-32B, and DeepSeek-R1) to generate reasoning processes. Code is extracted via regular expressions and executed through a Python interpreter (Appendix~\ref{Appd:answer_generation}). 
As shown in Figure~\ref{Fig:explation}, these models encompass both thinking and no-thinking modes. We first manually annotate the 5,523 <table, question, answer> triples (Appendix~\ref{Appd:answer_annotation_procedure}) to serve as the foundation for Reinforcement Learning (RL). Subsequently, we filter the reasoning traces generated by the 6 LLMs. For instances with multiple valid paths, we retain only the most representative high-quality process through manual election (Appendix~\ref{Appd:question_reasoning_annotaion_procedure}). This yields 2 specialized SFT datasets for thinking and no-thinking modes, each containing 1,932 <table, question, reasoning process> triples. To our knowledge, this is the first industrial-scale TableQA dataset featuring systematically annotated dual-mode reasoning paths.

\captionsetup[subfloat]{font=scriptsize}
\begin{figure}[t!]
    \captionsetup[subfigure]{captionskip=-10pt, nearskip=-5pt}
    \centering
	\begin{minipage}{0.65\linewidth}
        \centering 
        \subfloat[]{
        \includegraphics[width=\textwidth]{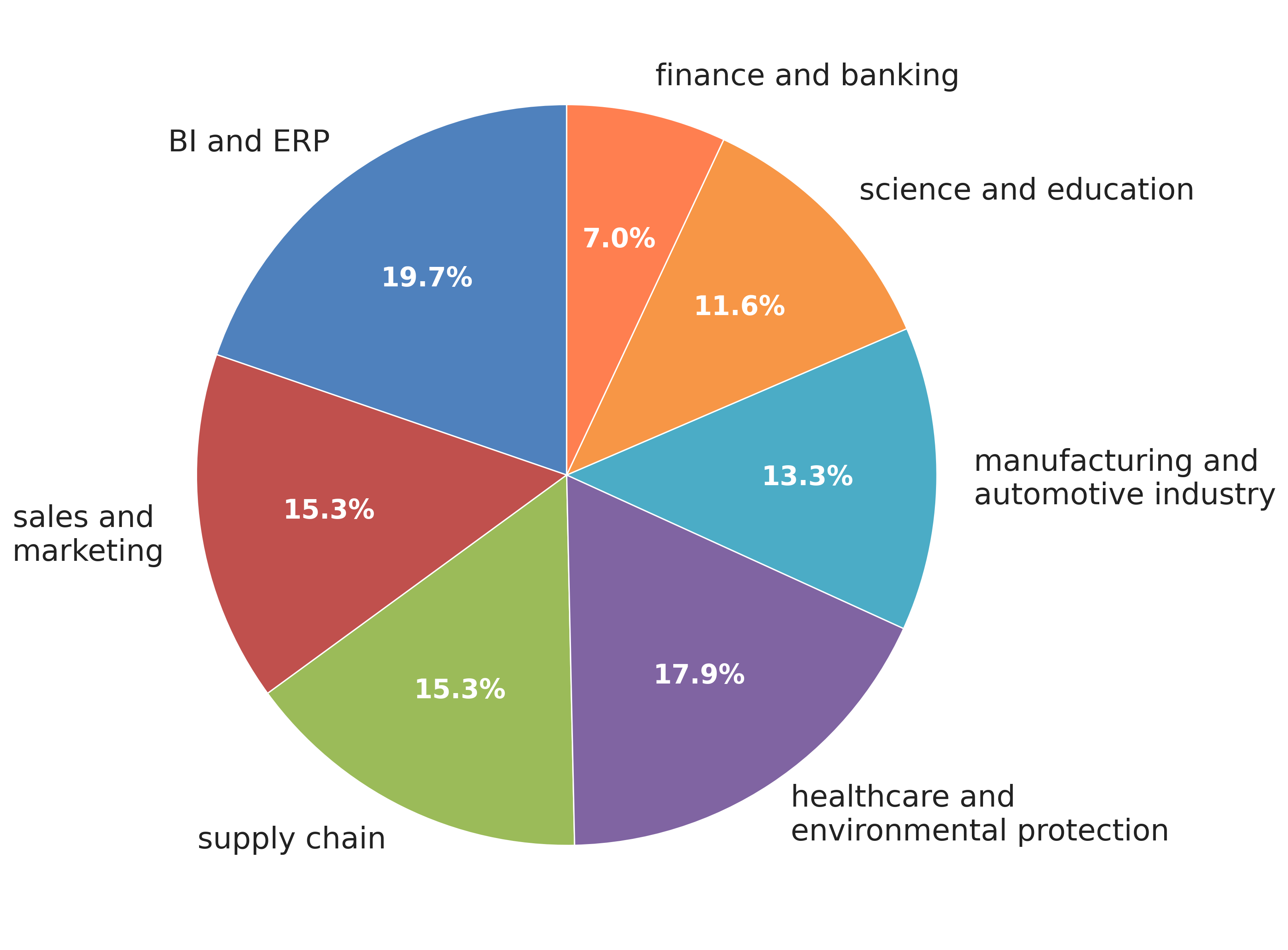}
        \label{Fig:domain_distribution}
        }
        \vspace{5pt} 
        \end{minipage}
    \captionsetup[subfigure]{captionskip=0pt, nearskip=0.0pt}
    \begin{minipage}{0.30\linewidth}
        \centering
	\begin{minipage}{\linewidth}
        \centering 
        \subfloat[]{
        \includegraphics[width=\textwidth]{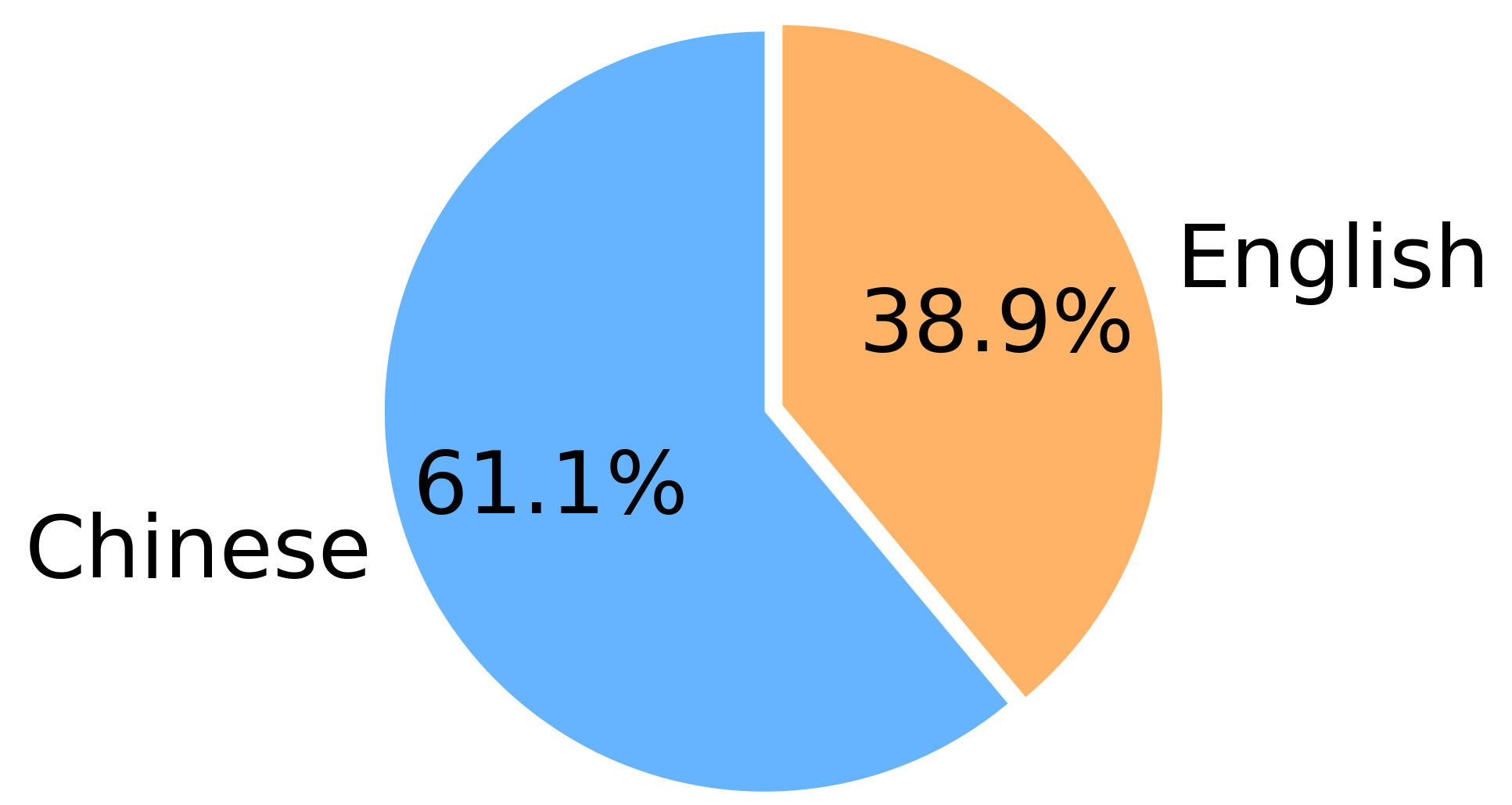} 
        }
        \end{minipage}
        \begin{minipage}{\linewidth}
        \centering 
        \subfloat[]{
        \includegraphics[width=\textwidth]{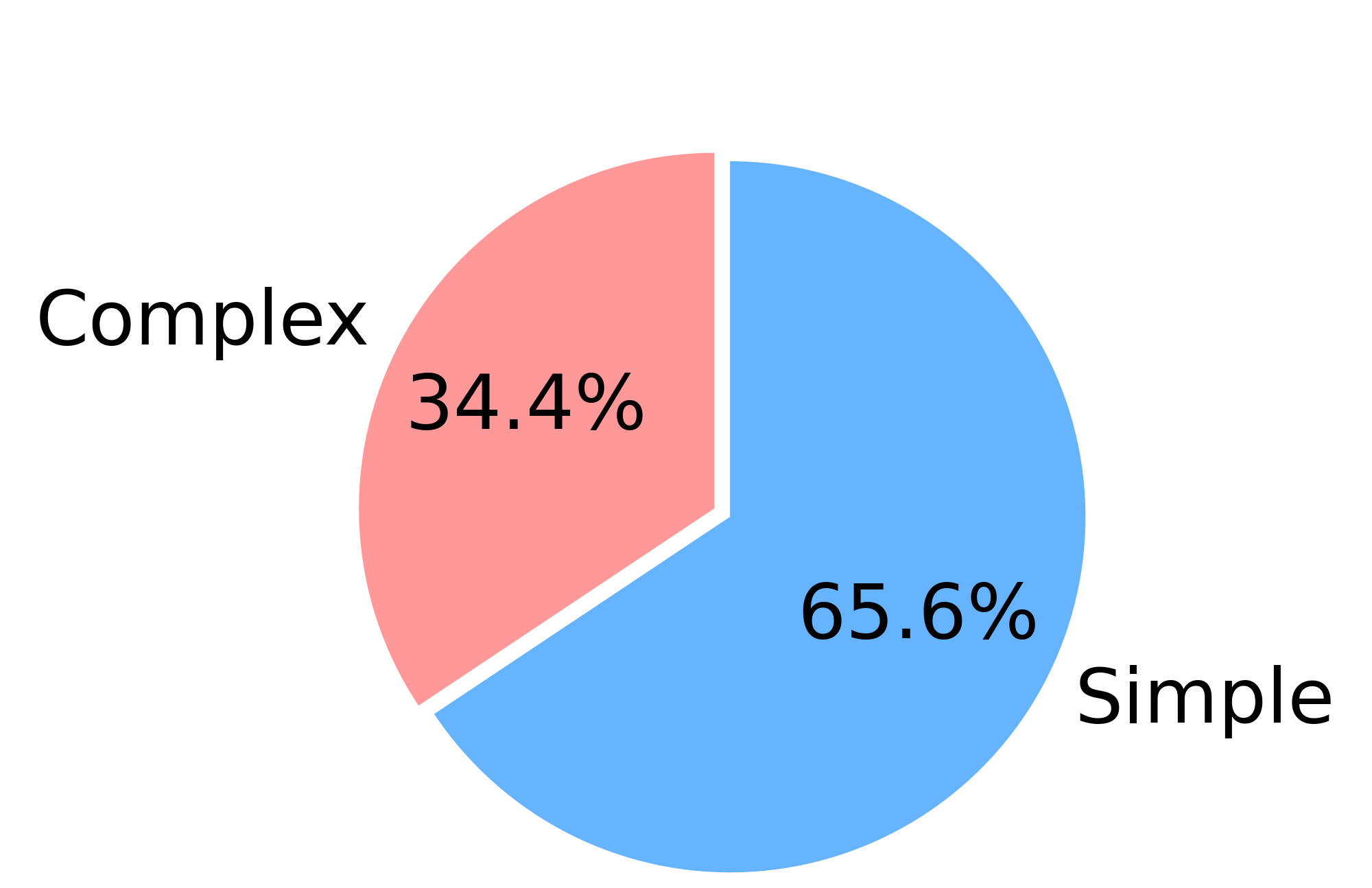} 
        }
        \end{minipage}
    \end{minipage}
        
        \begin{minipage}{0.48\linewidth}
        \centering 
        \subfloat[]{
        \includegraphics[width=\textwidth]{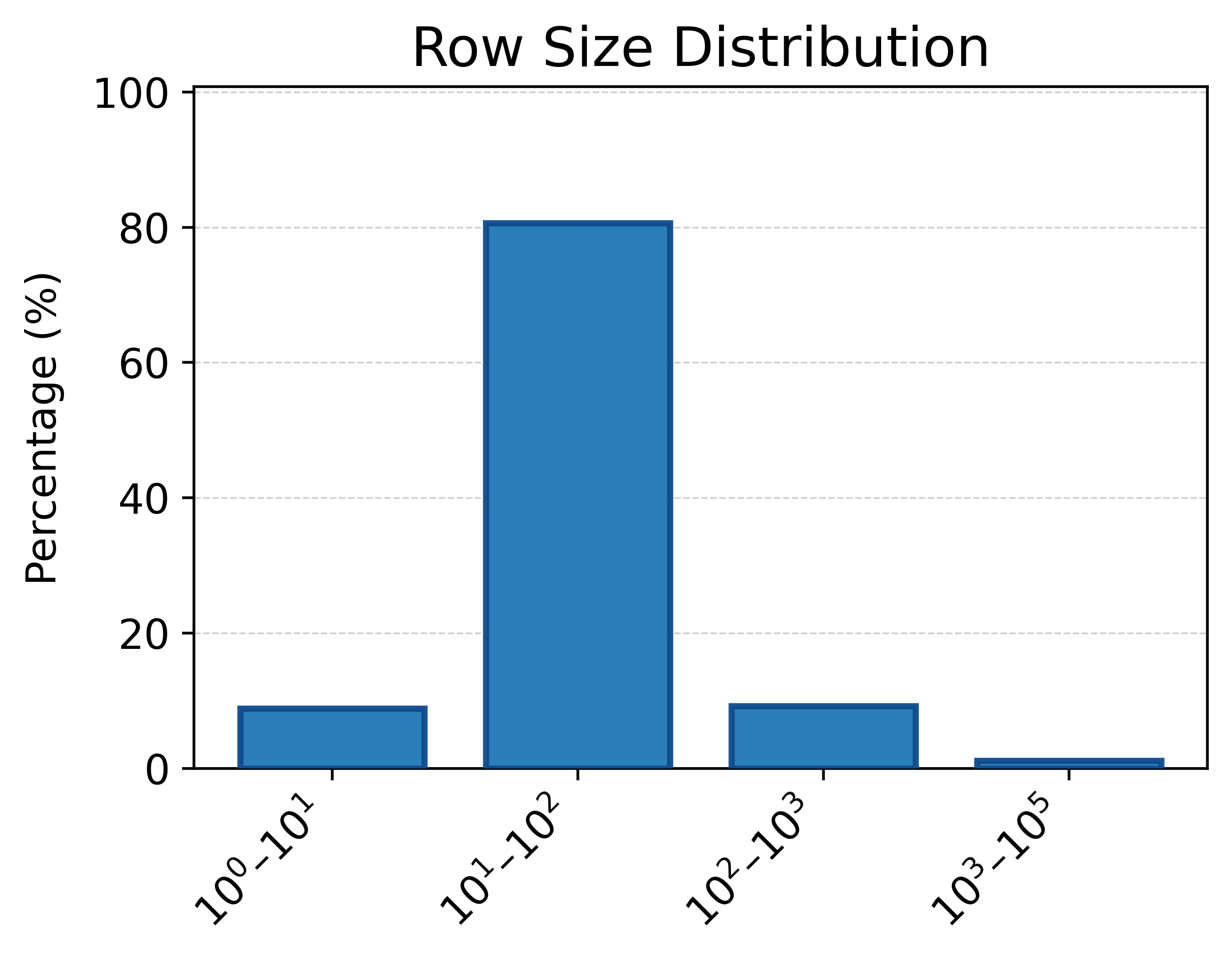} 
        }
        \end{minipage}
        \begin{minipage}{0.48\linewidth}
            \centering 
            \subfloat[]{
            \includegraphics[width=\textwidth]{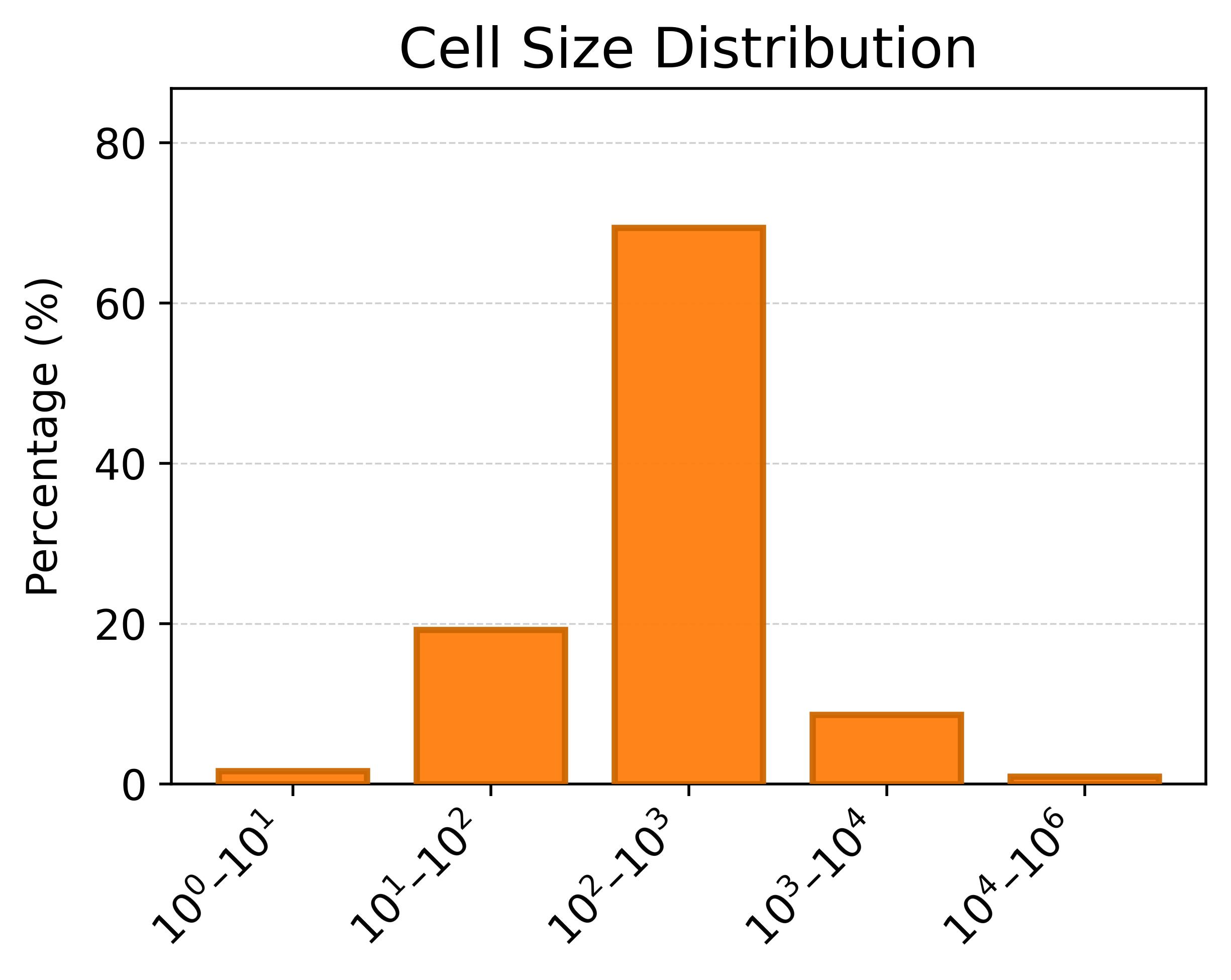} 
            }
            \end{minipage}
        
        \begin{minipage}{0.48\linewidth}
        \centering 
        \subfloat[]{
        \includegraphics[width=\textwidth]{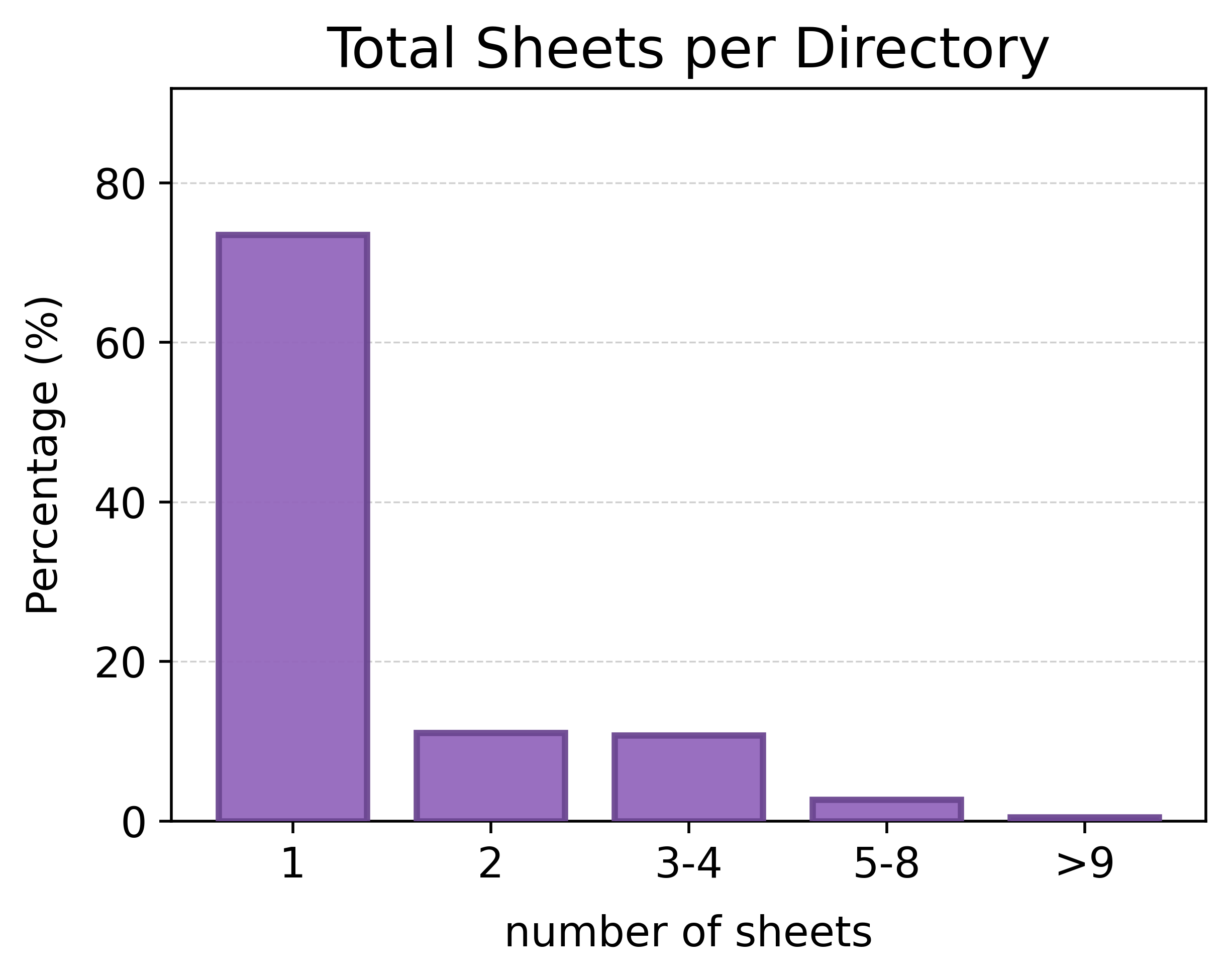} 
        }
        \end{minipage}
        \begin{minipage}{0.48\linewidth}
        \centering 
        \subfloat[]{
        \includegraphics[width=\textwidth]{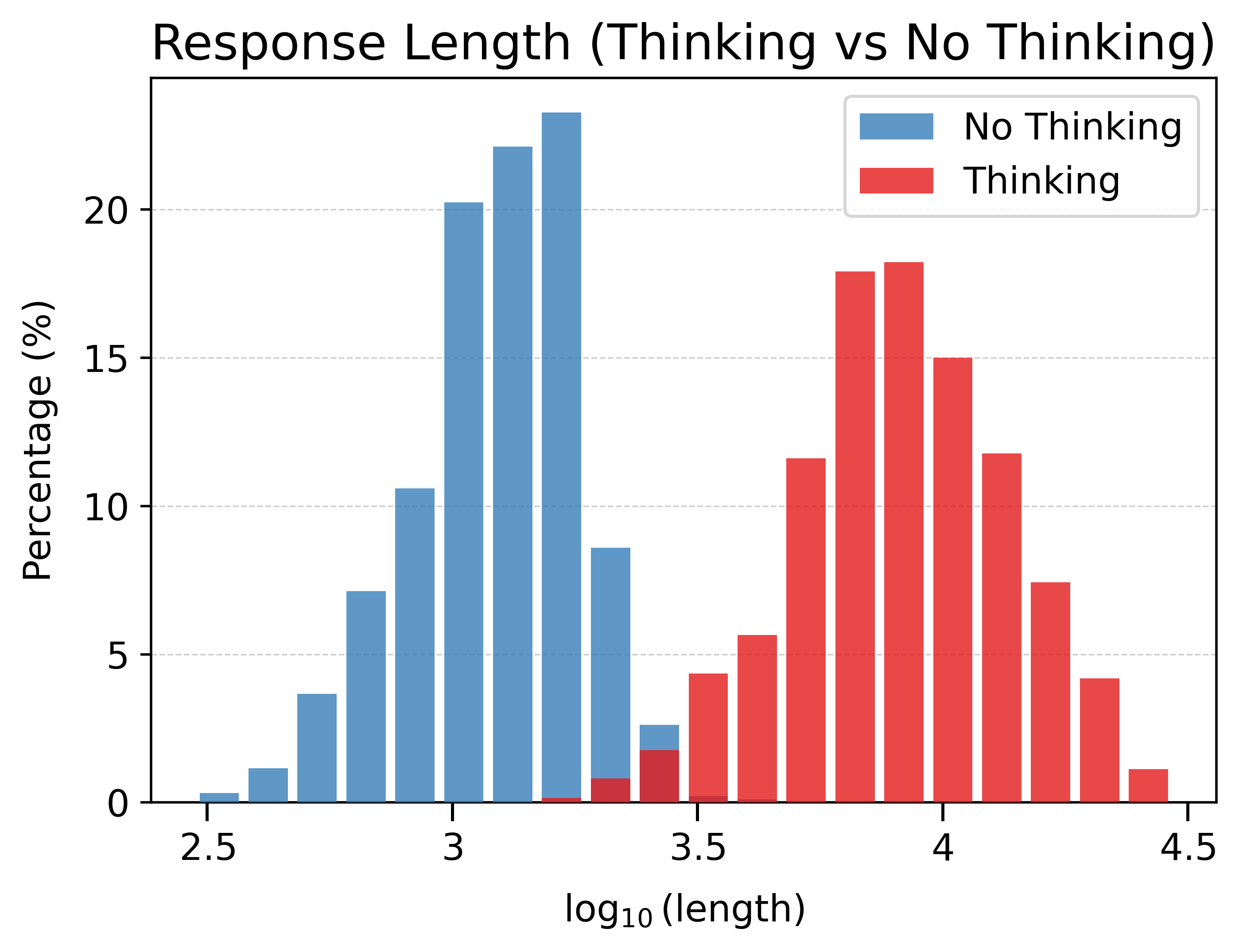} 
        \label{Fig:reasoing_modes_response_length}
        }
        \end{minipage}
\caption{Distribution of different types of tables in ReasonTabQA. (a) Domain distribution. (b) Proportion of Chinese and English tables. (c) Proportion of complex header tables. (d-e) The row and cell size distribution for all tables. (f) Proportion of sheets number in each directory. (g) Proportion of response length for SFT datasets with different reasoning modes.}
\label{Fig:data_distribution}
\end{figure}

\subsection{Dataset Statistics}
\label{Sec:dataset_statistics}

Through the construction process, ReasonTabQA comprises 5,523 high-quality questions originating from 1,932 unique tables. This collection includes 5,523 annotated <table, question, answer> triples and two SFT datasets, each containing 1,932 annotated <table, question, reasoning process> triples for thinking and no-thinking modes, respectively.

\paragraph{Structural Complexity.} Table~\ref{Tab:key_stat} and Figure~\ref{Fig:data_distribution} present the key statistics and structural distribution. Notably, ReasonTabQA is characterized by its high-fidelity industrial complexity: 8.3\% are extremely large-scale tables containing over 50K cells, 34.4\% feature complex structures such as hierarchical indexing and non-uniform merged cells, and 28.3\% are multi-table or multi-sheet configurations.

\paragraph{Domain Distribution.} As illustrated in Figure~\ref{Fig:domain_distribution}, the benchmark spans seven primary industrial domains, further categorized into 30 specialized sub-domains including marketing, manufacturing, automotive, business intelligence (BI), enterprise resource planning (ERP), and supply chain. Detailed sub-categories are provided in Table~\ref{Tab:domains} of Appendix~\ref{Appd:domain_distribution}, ensuring the dataset encapsulates the diversity of real-world industrial scenarios.

\paragraph{Response Characteristics.} A distinguishing feature of ReasonTabQA is the substantial variation in reasoning density across its two SFT datasets. As shown in Figure~\ref{Fig:reasoing_modes_response_length}, the response length for thinking mode (ranging from 1,867 to 29,639 tokens) is significantly more extensive than that of no-thinking mode (ranging from 298 to 3,823 tokens). This collection of dual-mode reasoning traces provides a unique resource for exploring logical elicitation and inference scaling in TableQA.

\begin{figure*}[t]
    \centering 
    \includegraphics[width=0.88\textwidth]{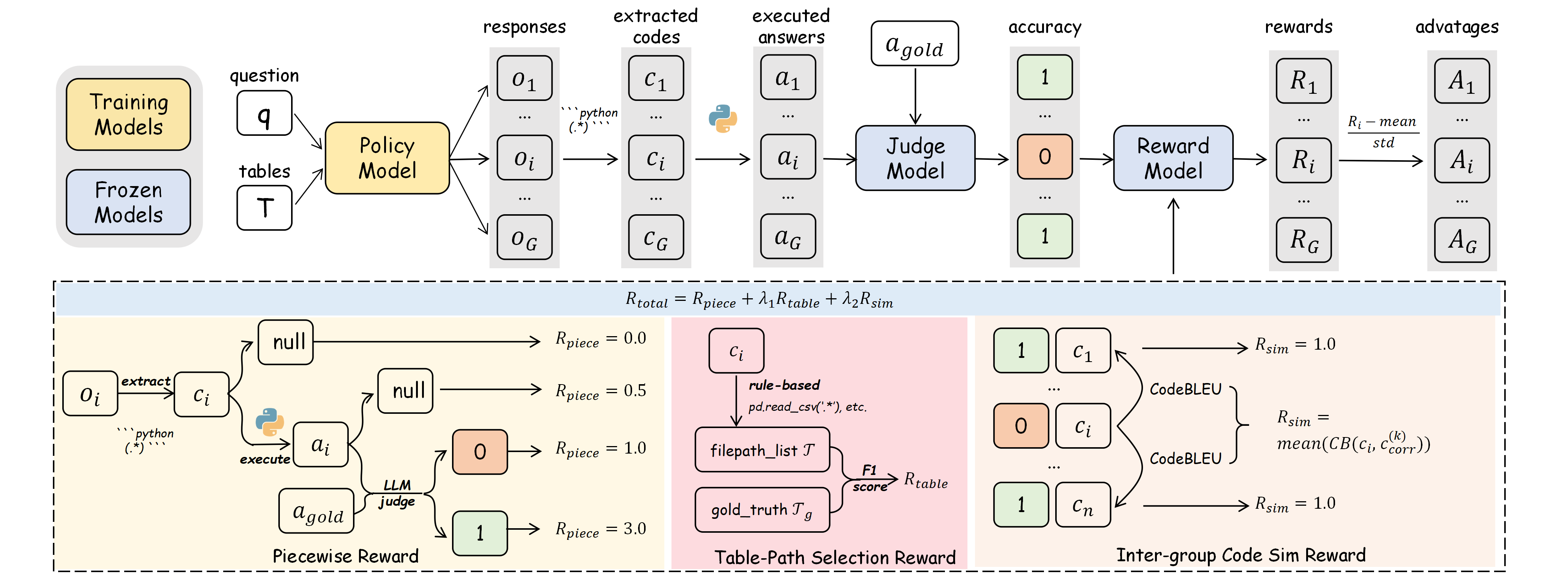} 
    \caption{Overview of TabCodeRL. The TabCodeRL method integrates piecewise discrete rewards with inner-group code semantic similarity rewards to provide granular optimization signals.}
    \label{Fig:method}
\end{figure*}

\section{Method}
\label{Sec:method}

To strengthen the model’s table reasoning capability on complex structural industrial tables, we propose TabCodeRL, as illustrated in Figure~\ref{Fig:method}. It is a two-stage training pipeline consisting of a cold-start phase and TabCodeRL optimization. It incorporates 3 components: multi-stage piecewise reward, table-path selection reward, and verifiable inner-group code similarity reward.  


\subsection{Cold Start}
For robust model initialization, we perform a cold-start Supervised Fine-Tuning (SFT) stage using ReasonTabQA. We differentiate the training data based on the target reasoning paradigm: no-thinking-mode data is assigned to non-reasoning models, while thinking-mode data is employed for reasoning models.


\subsection{RLVR Algorithm}
Based on the finetuned model, we apply DAPO \cite{yu2025dapoopensourcellmreinforcement} to train the reasoning model. Specifically, given table $T$ and question $q$, the policy $\pi(\theta)$ generates a group of $G$ candidate responses $\{o_i\}_{i=1}^G$. 
The objective function of DAPO is:
\begin{equation}
\small
\begin{aligned}
\mathcal{J}_{\text{DAPO}}(\theta) &= \mathbb{E}_{\substack{(q, a) \sim \mathcal{D} \\
\{o_i\}_{i=1}^G \sim \pi_{\theta_{\text{old}}}(\cdot|q)}} 
\Bigg[ 
\frac{1}{\sum_{i=1}^G |o_i|} 
\sum_{i=1}^G \sum_{t=1}^{|o_i|} \\
\min \Bigl(&
r_{i,t}(\theta) \hat{A}_{i,t}, 
\operatorname{clip}\bigl( 
r_{i,t}(\theta), 
\, 1 - \varepsilon_{\text{low}}, 
\, 1 + \varepsilon_{\text{high}} 
\bigr) \hat{A}_{i,t} 
\Bigr) 
\Bigg] \\
\text{s.t.} & \quad  0 < \left| \left\{ o_i \text{ is correct} \right\} \right| < G,
\end{aligned}
\end{equation}
where advantage of the $i$-th response at token step $t$, $\hat{A}_{i,t}$, is computed by group normalization:
\begin{equation}
\small
\hat{A}_{i,t} = \frac{R_i - \mu_{\mathcal{G}}}{\sigma_{\mathcal{G}}}, 
\end{equation}
with $\mu_{\mathcal{G}}$ and $\sigma_{\mathcal{G}}$ being the mean and standard deviation of $\{R_i\}_{i=1}^G$.

\subsection{Rewards Design}


\paragraph{Piecewise Reward.}

For each model-generated response $o_i$, let $c_i$ denote the executable Python code $c_i$ extracted from $o_i$, whose execution yields the final answer $a_{i}$, and the correctness is determined by comparing $a_{i}$ with the gold answer $a_{g}$. 
Based on these definitions, the correctness assessment of generated code output is decomposed into three sequential validation stages called format correctness, execution success and answer correctness, leading to a piecewise reward as follow:
\begin{equation}
\small
R_{\text{piece}} (o_i)= 
\begin{cases} 
0.0, & \text{ext}(o_i) = \emptyset \\ 
0.5, & \text{ext}(o_i) \neq \emptyset \land \text{exe}(c_i) = \emptyset \\ 
1.0, & \text{exe}(c_i) \neq \emptyset \land J(a_i, a_g) = 0 \\ 
3.0, & \text{exe}(c_i) \neq \emptyset \land J(a_i, a_g) = 1 
\end{cases}
\end{equation}
where $\text{ext}(o_i)=\emptyset$ denotes a format error when regex-based code extraction fails, the execution output $\text{exe}(c_i) =\emptyset$ denotes code execution failure, and $J(a_i, a_{g})$ represents a binary evaluator leveraging an LLM for correctness assessment.


\paragraph{Table-Path Selection Reward.}



During the code generation phase in industrial table QA, models often face challenges such as hallucinations caused by long and complex table paths, or errors in selecting the question-related tables from multiple candidates, resulting in straightforward code failure at the outset (see the Figure~\ref{Fig:case_study_error} for illustrations). To address this critical issue, we introduce a table-path selection reward, which is calculated by comparing the collections of table paths $\mathcal{T}$ extracted from the generated code $c_i$ by regex based methods, with the collection of ground-truth table paths $\mathcal{T}_g$ annotated for answering the question, and computing their F1-Score as below:
\begin{equation}
\small
R_{table}(o_i) = F1(\mathcal{T}, \mathcal{T}_g)
\end{equation}

\paragraph{Innergroup Code Similarity Reward.}


We introduce the inner-group code similarity reward to steer the code refinement process from incorrect to correct states. This reward calculates the CodeBLEU~\cite{ren2020codebleumethodautomaticevaluation} semantic similarity between incorrect but executable code $c_{err}$ and reference correct implementations $c_{corr}$. 
Unlike general code generation, table-based tasks (e.g., Pandas) exhibit uniform solution patterns, making them less prone to multi-solution ambiguity.
Inspired by GRPO~\cite{shao2024deepseekmathpushinglimitsmathematical}, we use the correct codes within the same group as reference codes. Since the incorrect and correct codes originate from the same group and share consistent style, the erroneous code can learn more effectively. The definition is as follows:
\begin{equation}
\small
R_{\text{sim}}(o_i) = 
\begin{cases} 
\frac{1}{n_{c}} \sum_{k=1}^{n_{c}} \text{CB}(c_{i}, c_{\text{corr}}^{(k)}), & \text{if } c_i \text{ is wrong} \\ 
1.0, & \text{if } c_i \text{ is correct} 
\end{cases}
\end{equation}
where $\text{CB}(.)$ means calculating the CodeBLEU score, $n_{c}$ is the number of correct implements in the group. Notably, since DAPO's dynamic sampling ensures each group contains at least one correct sample,  $n_{c}$ never equals zero in this computation.

The final reward is computed as the summation of the $R_{piece}(o_i)$, $R_{table}(o_i)$ and $R_{sim}(o_i)$, where the latter is scaled by hyperparameters $\lambda_1$ and $\lambda_2$ (with default values 0.5 and 1.0) to control its relative contribution: 
\begin{equation}
\small
R_{total}(o_i) = R_{piece}(o_i) + \lambda_1 R_{table}(o_i) + \lambda_2 R_{sim}(o_i)
\end{equation}






\section{Experiments}

\begin{table*}[!ht]
\small
\centering
\setlength{\tabcolsep}{4.2pt} 
\renewcommand{\arraystretch}{1.15} 

\resizebox{0.72\textwidth}{!}{%
\begin{tabular}{l|c|cc|ccc|ccc}
    \toprule
    \multirow{2}{*}{\textbf{Model}} & \multirow{2}{*}{\textbf{Overall}} & \multicolumn{2}{c|}{\textbf{Language}} & \multicolumn{3}{c|}{\textbf{Question Difficulty}} & \multicolumn{3}{c}{\textbf{Table Difficulty}} \\ 
    & & \textbf{English} & \textbf{Chinese} & \textbf{Easy} & \textbf{Medium} & \textbf{Hard} & \textbf{Simple} & \textbf{Medium} & \textbf{Complex} \\
    \midrule

    \rowcolor{SectionBg}
    \multicolumn{10}{c}{\textit{\textbf{No Reasoning Models}}} \\
    \midrule
    \multicolumn{10}{l}{\hspace{0.2cm}\textbf{Open-Source Models}} \\ 
    \midrule
    Qwen2-72B-Instruct \cite{yang2024qwen2} & 36.66 & 40.00 & 34.26 & 56.90 & 32.04 & 33.33 & 39.39 & 31.03 & 38.60 \\  
    Qwen2.5-Coder-32B-Instruct \cite{hui2024qwen25codertechnicalreport} & 45.28 & 47.10 & 43.98 & 58.62 & 37.86 & 45.24 & 50.00 & 36.21 & 47.37 \\ 
    Qwen2.5-72B-Instruct \cite{qwen2025qwen25technicalreport} & 51.92 & 53.55 & 50.75 & 61.42 & 52.32 & 49.10 & 55.10 & 50.98 & 42.79 \\ 
    Qwen3-8B (no-thinking) \cite{qwen3} & 40.97 & 43.23 & 39.35 & 52.07 & 41.55 & 37.62 & 44.95 & 40.48 & 28.14 \\ 
    Qwen3-32B (no-thinking) \cite{qwen3} & 53.13 & 49.03 & 56.07 & 57.21 & 52.79 & 52.17 & 56.85 & 51.12 & 44.30 \\ 
    Llama-3.1-8B-Instruct \cite{llama3} & 33.24 & 36.23 & 31.09 & 39.12 & 32.41 & 32.02 & 37.19 & 32.12 & 21.80 \\ 
    Llama-3.3-70B-Instruct \cite{llama3} & 37.74 & 38.10 & 37.48 & 46.55 & 39.81 & 34.29 & 38.89 & 40.52 & 28.07 \\ 
    Mistral-Large-Instruct-2407 \cite{jiang2023mistral} & 47.17 & 47.10 & 47.22 & 62.07 & 46.60 & 43.33 & 53.03 & 42.24 & 36.84 \\ 
    Telechat3-36B \cite{telechat-2024} & 51.13 & 53.51 & 49.42 & 58.49 & 50.98 & 49.17 & 54.12 & 51.99 & 38.99 \\ 
    Deepseek-V3 \cite{deepseekai2024deepseekv3technicalreport} & 52.15 & 54.81 & 50.24 & 57.12 & 53.66 & 50.04 & 55.07 & 51.12 & 44.10 \\ 
    Kimi-K2-Instruct \cite{kimiteam2025kimik15scalingreinforcement} & 59.57 & 54.19 & 63.43 & 62.43 & 60.09 & 58.52 & 63.13 & 56.17 & 54.11 \\ 
    \midrule
    \multicolumn{10}{l}{\hspace{0.2cm}\textbf{Closed-Source Models}} \\ 
    \midrule
    GPT-4o \cite{gpt4} & 56.90 & 54.25 & 58.80 & 60.90 & 59.01 & 54.76 & 63.15 & 51.69 & 45.79 \\ 
    GPT-5.2 & 59.46  & 61.29  & 58.14  & 64.79  & 61.25  & 57.11  & 66.29  & 53.81  & 47.23 \\ 
    \midrule
    \multicolumn{10}{l}{\hspace{0.2cm}\textbf{Table-Specific Models}} \\ 
    \midrule
    TableGPT2-7B \cite{su2024tablegpt2largemultimodalmodel} & 42.05 & 45.16 & 39.81 & 53.45 & 40.86 & 39.48 & 45.96 & 37.07 & 38.60 \\ 
    TableLLM-7B \cite{zhang2024tablellm} & 13.50 & 15.51 & 12.06 & 17.49 & 14.15 & 12.08 & 14.12 & 12.93 & 12.51 \\ 
    TableLlama-7B \cite{zhang2024tablellama} & 13.23 & 17.70 & 10.02 & 15.58 & 14.12 & 12.14 & 15.12 & 13.54 & 6.03 \\ 
    \midrule

    \rowcolor{ReasoningBg}
    \multicolumn{10}{c}{\textit{\textbf{Reasoning Models}}} \\
    \midrule
    \multicolumn{10}{l}{\hspace{0.2cm}\textbf{Open-Source Models}} \\ 
    \midrule
    QWQ-32B \cite{qwq32b} & 54.10 & 55.48 & 53.11 & 61.90 & 56.99 & 50.53 & 57.70 & 53.71 & 42.39 \\ 
    Qwen3-8B (thinking) \cite{qwen3} & 49.87 & 43.23 & 54.63 & 62.07 & 49.69 & 46.59 & 56.06 & 44.58 & 39.12 \\ 
    Qwen3-32B (thinking) \cite{qwen3} & 58.76 & 56.77 & 60.19 & 63.79 & 60.40 & 56.57 & 63.13 & 54.78 & 51.68 \\ 
    Qwen3-30B-A3B (thinking) \cite{qwen3} & 53.61 & 52.00 & 54.77 & 59.81 & 56.32 & 50.57 & 54.49 & 53.79 & 50.18 \\ 
    DeepSeek-R1-Dist-Qwen-7B \cite{deepseekai2025deepseekr1incentivizingreasoningcapability} & 11.86 & 16.13 & 8.80 & 13.90 & 11.68 & 11.39 & 17.68 & 6.03 & 3.51 \\ 
    DeepSeek-R1-Dist-Qwen-14B \cite{deepseekai2025deepseekr1incentivizingreasoningcapability} & 39.89 & 36.13 & 42.59 & 44.83 & 39.83 & 38.55 & 43.43 & 37.05 & 33.35 \\ 
    DeepSeek-R1-Dist-Qwen-32B \cite{deepseekai2025deepseekr1incentivizingreasoningcapability} & 53.64 & 54.84 & 52.78 & 58.45 & 55.09 & 51.60 & 58.59 & 48.40 & 47.12 \\ 
    Deepseek-R1 \cite{deepseekai2025deepseekr1incentivizingreasoningcapability} & 55.80 & 55.11 & 56.30 & 58.90 & 58.11 & 53.81 & 58.51 & 52.59 & 52.91 \\ 
    \midrule
    \multicolumn{10}{l}{\hspace{0.2cm}\textbf{Closed-Source Models}} \\ 
    \midrule
    Doubao-1.5-thinking-pro & 64.48 & 59.50 & \textbf{68.06} & 72.79 & \textbf{71.84} & 58.57 & 65.33 & 63.79 & \textbf{62.93} \\ 
    OpenAI o1-mini & 55.80 & 47.74 & 61.57 & 61.72 & 55.63 & 54.25 & 61.62 & 50.50 & 46.39 \\ 
    Claude-4.0-Sonnet & 62.53 & 60.00 & 64.35 & 70.69 & 61.35 & 60.86 & 64.14 & 61.56 & 58.91 \\ 
    Claude-Opus-4.5 & \underline{66.20}  & \underline{66.84}  & 65.74  & \underline{72.67}  & 69.18  & \underline{62.95}  & \underline{68.36}  & \underline{64.89}  & 61.36 \\ 
    Gemini-3-Pro-Preview & \textbf{67.58}  & \textbf{68.44}  & \underline{66.96}  & \textbf{74.64}  & \underline{69.27}  & \textbf{64.80}  & \textbf{70.15}  & \textbf{65.85}  & \underline{62.17} \\ 
    \midrule

    \rowcolor{OurMethodBg}
    \multicolumn{10}{c}{\textit{\textbf{TabCodeRL (ours)}}} \\
    \midrule
    TableLlama-7B-SFT-TabCodeRL & 20.43 & 23.83 & 17.99 & 21.92 & 21.26 & 19.61 & 22.71 & 19.15 & 15.11 \\
    DS-R1-Dist-Qwen-7B-SFT-TabCodeRL & 34.46 & 35.57 & 33.66 & 36.52 & 34.44 & 33.90 & 37.42 & 31.97 & 29.25 \\ 
    Qwen3-8B-NoThink-SFT-TabCodeRL & 58.01 & 54.22 & 60.73 & 61.59 & 59.40 & 56.34 & 60.69 & 56.92 & 50.92 \\ 
    Qwen3-8B-Think-SFT-TabCodeRL & \cellcolor{blue!8}{61.89} & \cellcolor{blue!8}{57.92} & \cellcolor{blue!8}{64.74} & \cellcolor{blue!8}{68.49} & \cellcolor{blue!8}{61.99} & \cellcolor{blue!8}{60.02} & \cellcolor{blue!8}{63.89} & \cellcolor{blue!8}{61.55} & \cellcolor{blue!8}{55.63} \\  
    \bottomrule
\end{tabular}%
}
\caption{Overall performance of LLMs on ReasonTabQA. \textbf{Bold}/\underline{underlined} fonts denote the best/second-best results, and results in \colorbox{blue!8}{purple} indicate the best results among open-sourced LLMs. 
}
\label{Tab:overall_performance}
\end{table*}

\subsection{Experimental Settings}
\label{Sec:exp_setting}

\paragraph{Baselines.}
To systematically assess ReasonTabQA, we perform experiments across 29 baselines, including: (1) open-source models TableGPT2~\citep{su2024tablegpt2largemultimodalmodel},
TableLLM~\cite{zhang2024tablellm}, TableLlama~\cite{zhang2024tablellama}, Qwen series ~\citep{qwen, qwen2, qwen2025qwen25technicalreport, hui2024qwen25codertechnicalreport}, Llama family ~\citep{llama3}, Mistral~\citep{jiang2023mistral}, Deepseek models \citep{deepseekai2024deepseekv3technicalreport,deepseekai2025deepseekr1incentivizingreasoningcapability}, Kimi-K2 and TeleChat~ \citep{telechat-2024, wang2025technicalreporttelechat2telechat25}, and (2) closed-source models GPT series \citep{gpt4}, OpenAI o1-mini, Gemini3, Claude and Doubao series.

\paragraph{Evaluation Metrics and Training Details.}
We adopt Accuracy as our primary metric, determined by strictly matching model outputs with gold-standard answers. TabReasonQA is evaluated across three core dimensions: linguistic diversity (English vs. Chinese), structural complexity (simple, medium, and complex tables), and reasoning intensity (easy, medium, and hard questions). We also perform a comparative analysis against established TableQA benchmarks, including WikiTQ \citep{WikiTableQuestion}, AITQA \citep{katsis2022ait}, MiMoTable \cite{li2024mimotable}, and HiTab \cite{HiTab}. To ensure a rigorous comparison, all models utilize a unified prompt template (see Appendix~\ref{Sec:reasoning_prompt}). We partition ReasonTabQA into training and test sets with an 8:2 ratio. The training set is further bifurcated into 2 equal subsets for SFT and RL, respectively. For RL training, we set the clipping hyperparameters $\epsilon_{\text{high}}=0.28$ and $\epsilon_{\text{low}}=0.2$. Refer to Appendix~\ref{sec:traning_details} for details.

\subsection{Main Results}


\paragraph{Overall Performance.}
As shown in Table~\ref{Tab:overall_performance}, we find: 
(1) leading closed-source models such as Gemini-3-Pro-Preview and Claude-Opus-4.5 achieve only suboptimal results (the overall results only reach 67.58\%), highlighting the intrinsic difficulty of industrial TableQA. Moreover, these models operate as black boxes with undisclosed model scales and post-processing strategies, which limits their reproducibility. 
(2) Table-specific models (e.g., TableLlama) exhibit notable limitations in program-based reasoning, likely due to a fine-tuning bias toward direct answer generation rather than executable code synthesis and complex reasoning structures. 
(3) Model performance shows high sensitivity to task complexity, with substantial degradation on hard questions ($-9.95\%$) and complex tables ($-10.06\%$), validating the effectiveness of ReasonTabQA’s difficulty stratification. 
(4) Models equipped with explicit reasoning mechanisms (``thinking'' models) consistently outperform non-reasoning variants by $5\%$--$10\%$, indicating that explicit reasoning chains are essential for structural understanding of tabular data. 
(5) The proposed TabCodeRL framework achieves $7\%$--$20\%$ absolute improvements over existing open-source LLMs; notably, TabCodeRL-enhanced Qwen3-8B-Instruct surpasses its 32B counterpart and GPT5.2, demonstrating that verifiable reward optimization enables compact models to rival substantially larger and more opaque architectures.

\vspace{-10pt}
\paragraph{Performance across Different Benchmarks.}
As summarized in Table~\ref{Tab:benchmark_compare}, we evaluate model performance on ReasonTabQA alongside 4 TableQA benchmarks. The results reveal a pronounced performance discrepancy, with average accuracy on ReasonTabQA being 10.4\% lower than on existing datasets. This substantial margin underscores the inherent complexity of our industrial-scale benchmark ReasonTabQA and highlights its utility in assessing model robustness under challenging scenarios, such as multi-sheet configurations, intricate table structures, and large-scale data environments.



\begin{figure*}[t!]
    \centering 
    \includegraphics[width=0.74\textwidth]{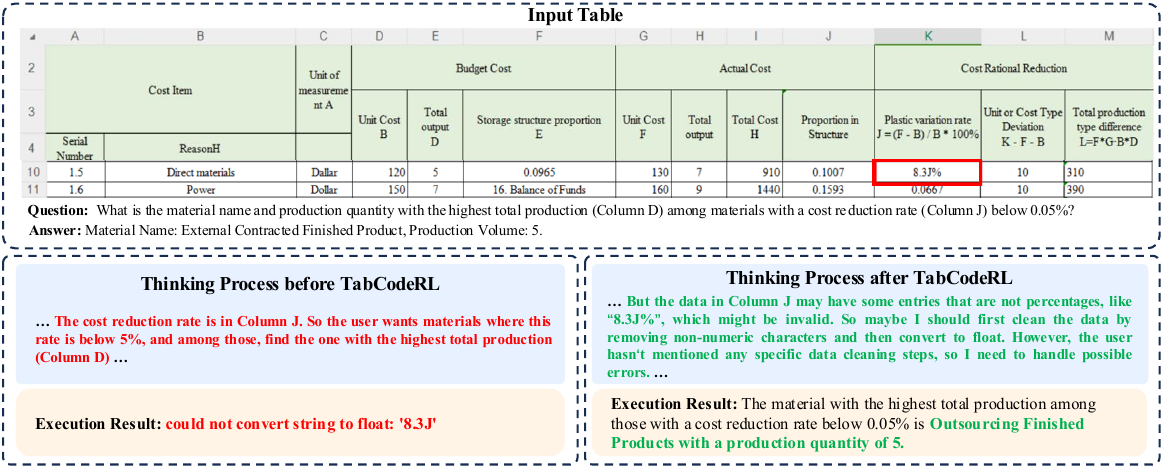} 
    \caption{Case study comparison of reasoning process before and after TabCodeRL.}
    \label{Fig:case_study_new}
\end{figure*}

\paragraph{Ablation Study.} 
We evaluate the contributions of SFT, DAPO, and TabCodeRL. As shown in Table~\ref{Tab:ablation_study}, we observe:
(1) TabCodeRL consistently outperforms both standalone SFT and DAPO by 0.1\%--2.4\%. This underscores that while SFT facilitates knowledge internalization, our table-specific verifiable rewards (RLVR) are essential for steering models toward generating executable and logically sound code.
(2) While SFT's improvements are primarily confined to in-domain distributions, RL-enhanced models exhibit a significant generalization dividend across four out-of-distribution benchmarks (WTQ, AITQA, MimoTable, HiTab). This robustness confirms that TabCodeRL captures transferable structural logic rather than relying on surface-level pattern matching.

\begin{table}[!t]
\small
\centering
\setlength{\tabcolsep}{5pt} 
\renewcommand{\arraystretch}{1.18} 

\resizebox{0.85\linewidth}{!}{%
\begin{tabular}{l|ccccc}
    \toprule
    \textbf{Model} & \textbf{ReasonTabQA} & \textbf{WTQ} & \textbf{AITQA} & \textbf{MimoTable} & \textbf{HiTab} \\ 
    \midrule

    \rowcolor{SectionBg}
    \multicolumn{6}{c}{\textit{\textbf{No reasoning Models}}} \\
    \midrule
    Qwen2.5-72B-Instruct & 51.92 & 77.45 & 58.59 & 67.28 & 81.03 \\ 
    Qwen3-8B-Instruct (no-thinking) & 40.97 & 66.03 & 61.55 & 51.57 & 61.62 \\
    Llama-3.1-8B-Instruct & 33.24 & 55.90 & 44.80 & 41.22 & 57.21 \\ 
    GPT-4o & 56.90 & 85.64 & 70.53 & 69.21 & 83.55 \\
    \midrule
    \rowcolor{ReasoningBg}
    \multicolumn{6}{c}{\textit{\textbf{Reasoning Models}}} \\
    \midrule
    QWQ-32B & 54.10 & 84.99 & 66.41 & 68.49 & 82.11 \\ 
    Qwen3-8B-Instruct (thinking) & 49.87 & 76.94 & 66.21 & 64.17 & 75.37 \\ 
    Qwen3-32B-Instruct (thinking) & 58.76 & 85.62 &  72.04 & 71.17 & 85.57 \\ 
    Deepseek-R1-Distill-Qwen-7B & 11.86 & 47.39 & 18.06 & 31.28 & 47.78 \\ 
    Deepseek-R1 & 55.80 & 83.92 & 69.05 & \textbf{72.86} & 82.99 \\ 
    OpenAI o1-mini & 55.80 & 80.57 & 66.89 & 66.74 & 87.11 \\ 
    Doubao-1.5-thinking-pro & 64.48 & 88.25 & 71.19 & \underline{72.45} & \underline{87.28} \\
    Claude-Opus-4.5 & \underline{66.20} & \underline{89.91} & \underline{72.22} & 70.04 & 86.45 \\  
    Gemini-3-Pro-Preview & \textbf{67.58} & \textbf{91.25} & \textbf{74.19} & 71.84 & \textbf{88.35} \\  
    \midrule

    \rowcolor{OurMethodBg} 
    \multicolumn{6}{c}{\textit{\textbf{Table-Specific Models}}} \\
    \midrule
    TableGPT2-7B & 42.05 & 62.15 & 49.89 & 41.56 & 48.87 \\ 
    TableLLM-7B & 13.50 & 31.74 & 21.32 & 15.14 & 15.90 \\ 
    TableLlama-7B & 13.23 & 25.65 & 11.33 & 21.49 & 23.68 \\ 
    \midrule
    Qwen3-8B-Think-SFT-TabCodeRL & 61.89 & 83.07 & 71.06 & 70.33 & 80.72 \\ 
    
    \bottomrule
\end{tabular}}
\caption{Performance comparison of different models on multiple TableQA benchmarks.}
\label{Tab:benchmark_compare}
\end{table}

\begin{table}[!t]
\small
\centering
\setlength{\tabcolsep}{4.5pt} 
\renewcommand{\arraystretch}{1.15} 

\resizebox{0.85\linewidth}{!}{%
\begin{tabular}{l|ccccc}
\toprule
\textbf{Setting} & \textbf{ReasonTabQA} & \textbf{WTQ} & \textbf{AITQA} & \textbf{MimoTable} & \textbf{HiTab} \\
\midrule

\rowcolor{SectionBg}
\multicolumn{6}{l}{\textit{Base: Qwen3-8B-Instruct (no-thinking)}} \\
\midrule
Vanilla Model & 40.97 & 66.03 & 61.55 & 51.57 & 61.62 \\ 
\quad + SFT & 53.40 & 67.50 & 62.72 & 53.50 & 62.47 \\ 
\quad + DAPO & 55.14 & 75.41 & 63.27 & 57.06 & 67.45 \\ 
\quad + TabCodeRL & 57.51 & 77.41 & 63.60 & 58.77 & 69.00 \\ 
\rowcolor{OurMethodBg}
\quad + SFT + TabCodeRL & 58.01 & 78.35 & 64.37 & 59.46 & 69.56 \\ 
\midrule

\rowcolor{SectionBg}
\multicolumn{6}{l}{\textit{Base: Qwen3-8B-Instruct (thinking)}} \\
\midrule
Vanilla Model & 49.87 & 76.94 & 66.21 & 64.17 & 75.37 \\ 
\quad + SFT & 58.90 & 79.04 & 69.06 & 64.84 & 77.52 \\ 
\quad + DAPO & 59.54 & 81.27 & 71.06 & 69.31 & 79.97 \\ 
\quad + TabCodeRL & 60.82 & 81.92 & 71.63 & 69.63 & 80.55 \\ 
\rowcolor{OurMethodBg}
\quad + SFT + TabCodeRL & 61.89 & 83.07 & 71.06 & 70.33 & 80.72 \\ 
\midrule

\rowcolor{SectionBg}
\multicolumn{6}{l}{\textit{Base: DeepSeek-R1-Distill-Qwen-7B}} \\
\midrule
Vanilla Model & 11.86 & 47.39 & 18.06 & 31.28 & 47.78 \\ 
\quad + SFT & 21.94 & 49.65 & 23.49 & 33.93 & 49.70 \\ 
\quad + DAPO & 31.86 & 53.09 & 28.84 & 37.36 & 54.42 \\ 
\quad + TabCodeRL & 33.42 & 54.76 & 28.44 & 38.00 & 56.21 \\ 
\rowcolor{OurMethodBg}
\quad + SFT + TabCodeRL & 34.46 & 58.09 & 32.37 & 40.85 & 57.02 \\ 

\bottomrule
\end{tabular}%
}
\caption{Ablation results on 5 TableQA benchmarks.}
\label{Tab:ablation_study}
\end{table}

\paragraph{Case Study.} 
We conduct a qualitative analysis to evaluate model performance on complex industrial scenarios. As shown in Figure~\ref{Fig:case_study_new}, TabCodeRL-enhanced Qwen3-8B-Instruct demonstrates significantly improved robustness. It accurately interprets complex structures and proactively identifies anomalous cells by generating defensive code, such as utilizing try-except blocks and error-handling pandas operations to mitigate runtime failures.  However, as shown in Figures~\ref{Fig:case_study_filepath}, \ref{Fig:case_study_error} and \ref{Fig:case_study_rl}, even the leading Gemini-3-Pro-Preview model occasionally fails to execute code due to a lack of contextual understanding of intricate table layouts. Detailed analysis are provided in Appendix~\ref{Appd:case_study}.



\section{Conclusion}
To bridge the gap in real-world industrial table reasoning, we present ReasonTabQA, an bilingual benchmark of 1,932 tables across 30 domains, featuring complex headers, multi-table structures, and large-scale data with dual-mode reasoning annotations (thinking and no-thinking).
Furthermore, we propose TabCodeRL, a reinforcement learning method that enhances reasoning capabilities through table-specific verifiable rewards. Extensive evaluations of 29 LLMs on ReasonTabQA and 4 established datasets demonstrate that while TabCodeRL yields substantial performance gains, the persistent performance margin on ReasonTabQA highlights the unique challenges of real world industrial TableQA.


\section*{Limitations}
While ReasonTabQA spans 30 diverse domains and represents the most extensive categorical coverage among existing TableQA benchmarks, it does not yet encapsulate the full spectrum of the global industrial landscape. Expanding this taxonomic breadth to include even more specialized sectors remains a key objective for future iterations. Additionally, although the benchmark establishes a robust bilingual foundation in Chinese and English, it lacks representation for other major languages with distinct morphological or syntactic structures, such as Arabic. We plan to broaden both the industrial scope and linguistic diversity of the dataset in future work to foster more universal and inclusive table reasoning systems.

\section*{Ethics Statement}
This paper studies LLMs using publicly available pretrained models and a benchmark curated for this work. The benchmark consists of tabular data collected from two primary sources: (1) public repositories, including municipal open data platforms, national statistical bureaus, and industry portals; and (2) industrial reports comprising anonymized real-world data and professional service datasets. All data are either publicly released or provided in anonymized form, and have undergone careful curation and filtering to remove private user data and personally identifiable information. The models are evaluated as-is, without any additional training or fine-tuning that could amplify harmful behaviors. We therefore believe that this study complies with the ACL Ethics Policy.

\bibliography{custom}

\appendix

\section*{Appendix}
\label{sec:appendix}

\section{Training and Evaluation Details}
\label{sec:traning_details}
We partition our dataset into an 8:2 ratio for training and test sets respectively. Within the training set, we further randomly allocate half of the samples for SFT and the remaining half for RL training. Leveraging the dual annotated reasoning process (thinking and non-thinking), we conduct SFT on specialized model variants for each cognitive processing paradigm. We adapt our method on the veRL framework\citep{sheng2024hybridflow} and follow the training recipe of DAPO for fair comparison: we use the clip ratio of $\epsilon_{high} = 0.28$ and $\epsilon_{low} = 0.2$. To accommodate lengthy tabular inputs, we extended both the prompt length and maximum response length to 16,384 tokens, ensuring comprehensive coverage of complex table structures while maintaining computational feasibility. All experiments are conducted on 4 nodes, each equipped with 8 $\times$ NVIDIA A800 80GB GPUs.

Due to flexible and open-ended nature of responses in our task, deterministic evaluation metrics like ROUGE-L or BLEU tend to yield underestimated scores. ROUGE-L and BLEU are two of the most classic n-gram–matching metrics in NLP. Both rely on counting overlapping words or phrases, so they inherently favor deterministic, short answers with a single reference. Grounded in real-world industrial scenarios, our dataset contains complex, multi-table, and large-scale tables. Consequently, our datasets answers often include not only concrete values but also analytical insights, making them inherently open-ended and flexible. ROUGE-L and BLEU alone are therefore inadequate for fully evaluating our datasets. Therefore, we employed solely the LLM-as-a-judge method, determining accuracy by comparing the model's outputs against the gold standard. 

\section{Implementation Details for Benchmark Construction}
\subsection{Data Source of ReasonTabQA}
\label{Appd:source}

The tables of bench are collected from publicly available internet resources and commercially purchased datasets. The internet sources include municipal open data platforms, the official website of the National Bureau of Statistics, industry association portals, and open-source tabular datasets, with data sources shown in Table~\ref{Tab:source}. 
Purchased data consists of various open-licensed tables and industrial reports acquired from professional data service providers, while our proprietary collection comprises anonymized tables accumulated from practical industrial applications.

\begin{table}[ht]
    \centering
    \resizebox{\linewidth}{!}{
    \begin{tabular}{l l}
    \toprule
     \textbf{Sources} & \textbf{Websites} \\
     \midrule
     \multicolumn{2}{l}{\textbf{Open-source data platform}} \\
     \midrule
     Wolrd Bank Group & https://datacatalog.worldbank.org/  \\
     \midrule
     National Bureau of Statistics of China & https://www.stats.gov.cn/sj/ \\ 
     \midrule
     Kaggle & https://www.kaggle.com/datasets  \\
     \midrule
     \makecell[l]{China Association of Automobile\\ Manufactures} & http://www.caam.org.cn/  \\
     \midrule
     Beijing Public Data Open Platform & https://data.beijing.gov.cn/  \\
     \midrule
     \makecell[l]{The United States Government’s \\Open Data Site} & https://catalog.data.gov/dataset  \\
     \midrule
     \makecell[l]{China Securities Regulatory \\Commission Data Platform} & http://www.csrc.gov.cn/csrc/tjsj/index.shtml  \\
     \midrule
     Shanghai Public Data Open Platform & https://data.sh.gov.cn/view/data-resource/index.html  \\
     \midrule
     CelesTrak & https://celestrak.org/  \\
     \midrule
     \multicolumn{2}{l}{\textbf{Tabular dataset}} \\
     \midrule
     MiMoTable\cite{li2024mimotable} & https://github.com/jasonNLP/MiMoTable  \\
     \bottomrule
    \end{tabular}}
    \caption{The data sources of ReasonTabQA Tables}
    \label{Tab:source}
\end{table}

\subsection{Details of Table and Question Difficulty Classification Criteria}
\label{Appd:question_table_difficulty_classification}
We design three difficulty levels to obviously distinct both the questions and table structures, yielding nine different difficulty types in total. 
Specifically, for question difficulty, as defined in Table~\ref{tab:question-difficulty}, we propose three prompt templates for each difficulty level (see Appendix~\ref{Appd:prompts_library_and_seed_questions_for_question generation}) to generate question of each difficulty seperately. 
For table structure difficulty, we propose three dimensions (multi-table, multi-sheet, complex headers) to two annotators for independent annotation, as shown in Table~\ref{tab:table-difficulty}. 

\begin{table}[!ht]
    \centering
    \resizebox{\linewidth}{!}{
    \begin{tabular}{c|l}
    \toprule
    \textbf{Difficulty} & \textbf{Defination} \\ 
    \midrule
    Easy & \makecell[l]{The question can be answered directly by retrieving \\values from the table without requiring any computation\\ or filtering.} \\ 
    \midrule
    Medium & \makecell[l]{The question requires simple computation, filtering, \\or conditional matching, typically solvable with\\ a single-step operation.} \\ 
    \midrule
    Hard & \makecell[l]{The question involves more than two steps of\\ reasoning, complex calculations, or comprehensive \\analysis, potentially including multi-condition filtering,\\ cross-table integration, and logical inference.} \\
    \bottomrule
    \end{tabular}}
    \caption{Question difficulty levels.}
    \label{tab:question-difficulty}
\end{table}

\begin{table}[h]
\centering
\small
\resizebox{\linewidth}{!}{
\begin{tabular}{c c c | c}
\toprule
\textbf{Multi-Table} & \textbf{Multi-Sheet} & \textbf{Complex Header} & \textbf{Difficulty} \\
\midrule
\ding{55} & \ding{55} & \ding{55} & Easy \\
\midrule
\ding{55} & \ding{51} & \ding{55} & Medium \\
\ding{55} & \ding{55} & \ding{51} & Medium \\
\midrule
\ding{55} & \ding{51} & \ding{51} & Hard \\
\ding{51} & any       & any       & Hard \\
\bottomrule
\end{tabular}}
\caption{Table structure difficulty levels. \ding{51} = Yes, \ding{55} = No, \textit{any} = either Yes or No. Multi-Table are always considered hard.}
\label{tab:table-difficulty}
\end{table}

\subsection{Details for Annotation Team Composition}
\label{Appd:annotation}

Our annotation team comprise 20 annotators, all possessing bilingual proficiency in English and Chinese demonstrated by standardized test scores such as IELTS 6.0, CET-6 or equivalent qualifications, alongside native fluency in Chinese. Each annotator hold a bachelor or higher degree and has at least one year of experience in data analysis, ensuring their capability to achieve high-quality annotations. The team covers all seven domains in ReasonTabQA, including eight senior annotators (holding master’s degrees in relevant fields) and twelve junior annotators. 
These senior members serve as quality control reviewers of questions relevant to their own domain, conducting final verification of annotations to ensure accuracy and consistency throughout the dataset development process.



All annotators work eight hours a day and earned a wage of \$40 per day on average. All annotators are trained through videos or online meetings provided with annotation guidelines that explains the data usage for academic research purposes.

\subsection{Prompts Library and Seed Questions for Question Generation}
\label{Appd:prompts_library_and_seed_questions_for_question generation}

\subsubsection{Easy Question Prompt}

The three prompt templates in the prompt library for easy question generation are shown below:
\begin{tcolorbox}[sidebyside, sidebyside align=top seam, width=\linewidth, colback=gray!20, colframe=white, colbacktitle=white, coltitle=white, breakable, arc=0mm, left=0mm, right=0mm]
\small

You are a professional expert in question generation based on tabular data, with expertise in designing questions grounded in table content. Below are 3 seed questions related to a table, along with the table description. Based on the analytical complexity and the reasoning pattern of the seed questions, please generate 3 new questions. Make sure that each question can be answered based solely on the provided table. Please read the following carefully:\\

\#\# Seed Questions:\\
{\red [SEED QUESTION 1]}\\
{\red [SEED QUESTION 2]}\\
{\red [SEED QUESTION 3]}\\

\#\# Tabular Data:\\
{\red [TABLE DESCRIPTION]}\\

\#\# Task Requirements:\\
\hspace{15pt}

Please refer to both the tabular data and the seed question to generate 3 new questions. Ensure that the generated questions meet the following criteria:\\
1.	Each question must be directly related to the provided tabular data and answerable based on that data.\\
2.	The questions should be simple, involving only one main subject and requiring just a single-step calculation or retrieval to answer.\\
3.	Each question must be semantically complete, clearly worded, and consistent with natural human expression.\\
4.	The questions should not be exact repetitions or too similar to the seed questions—ensure novelty and diversity in phrasing.\\
5.	The three questions must not be similar to each other; the questioning strategies should be flexible and varied.\\

\#\# Output Format:\\

\hspace{15pt}
\begin{minipage}{\dimexpr\linewidth-15pt}
Question 1:...\\
Question 2:...\\
Question 3:...\\
\end{minipage}

Do not include any other content in the output.

\end{tcolorbox}

\begin{tcolorbox}[sidebyside, sidebyside align=top seam, width=\linewidth, colback=gray!20, colframe=white, colbacktitle=white, coltitle=white, breakable, arc=0mm, left=0mm, right=0mm]
\small

You are a domain expert in designing table-based questions for data comprehension and analysis. Provided below are 3 sample (seed) questions and a textual description of a data table. Your task is to create 3 new questions that align in reasoning level and format with the seed questions, while introducing novel content and structure. \\

\#\# Seed Questions:\\
{\red [SEED QUESTION 1]}\\
{\red [SEED QUESTION 2]}\\
{\red [SEED QUESTION 3]}\\

\#\# Table Description:\\
{\red [TABLE DESCRIPTION]}\\

\#\# Instructions:\\
\hspace{15pt}

Use both the table context and the sample questions to inspire your generation. Ensure that the new questions follow these rules:\\
1.	Each question should rely strictly on the information present in the table description.\\
2.	Avoid multi-step reasoning—keep the logic shallow, using only one operation like filtering, lookup, or basic calculation.\\
3.	Ensure the questions are clear, natural, and precise in language, without ambiguity.\\
4.	The new questions should not simply restate or paraphrase the seed questions. Strive for originality in both content and style.\\
5.	The three questions should each focus on a different fact or angle from the table.\\

\#\# Output Format:\\

\hspace{15pt}
\begin{minipage}{\dimexpr\linewidth-15pt}
Question 1:...\\
Question 2:...\\
Question 3:...\\
\end{minipage}

Only include the questions—do not provide any explanations or commentary.

\end{tcolorbox}

\begin{tcolorbox}[sidebyside, sidebyside align=top seam, width=\linewidth, colback=gray!20, colframe=white, colbacktitle=white, coltitle=white, breakable, arc=0mm, left=0mm, right=0mm]
\small

You are a skilled question generation expert specialized in tabular reasoning. Below are three reference questions and a corresponding table description. Your goal is to create 3 new and distinct questions inspired by the style and complexity of the reference items.\\

\#\# Reference Questions:\\
{\red [SEED QUESTION 1]}\\
{\red [SEED QUESTION 2]}\\
{\red [SEED QUESTION 3]}\\

\#\# Table Context:\\
{\red [TABLE DESCRIPTION]}\\

\#\# Guidelines:\\
\hspace{15pt}

Use the reference questions and the table description to craft 3 original, answerable questions that meet the following criteria:\\
1.	Questions must be grounded solely in the table—no external knowledge should be required.\\
2.	Keep each question simple, focusing on direct retrieval or one-step aggregation.\\
3.	Language should be fluent, concise, and free of awkward constructions.\\
4.	Do not echo the seed questions in structure or focus—diversify your approach.\\
5.	Each question should target a different element, attribute, or data entry in the table.\\

\#\# Output Format:\\

\hspace{15pt}
\begin{minipage}{\dimexpr\linewidth-15pt}
Question 1:...\\
Question 2:...\\
Question 3:...\\
\end{minipage}

Only output the questions. Do not add any headings, explanations, or formatting notes.

\end{tcolorbox}

\noindent The 5 Seed Questions are shown below:

\begin{tcolorbox}[sidebyside, sidebyside align=top seam, width=\linewidth, colback=gray!20, colframe=white, colbacktitle=white, coltitle=white, breakable, arc=0mm, left=0mm, right=0mm]
\small

Seed Question 1: What is the total amount of the in-station cable project? \\

Seed Question 2: What was the corporate income tax revenue in 2021? \\

Seed Question 3: What was Hu Lianjie’s score in the chemistry subject? \\

Seed Question 4: What was Shandong Province’s expenditure on agricultural support in 1980 (in ten thousand yuan)? \\

Seed Question 5: What was the industrial added value of the non-ferrous metal mining and dressing industry in 2006 (in 100 million yuan)?
\end{tcolorbox}

\subsubsection{Medium Question Prompt}

The three prompt templates in the prompt library for medium question generation are shown below:

\begin{tcolorbox}[sidebyside, sidebyside align=top seam, width=\linewidth, colback=gray!20, colframe=white, colbacktitle=white, coltitle=white, breakable, arc=0mm, left=0mm, right=0mm]
\small

You are a professional expert in question generation based on tabular data, with expertise in designing questions grounded in table content. Below are 3 seed questions related to a table, along with the table description. Based on the analytical complexity and the reasoning pattern of the seed questions, please generate 3 new questions. Make sure that each question can be answered based solely on the provided table. Please read the following carefully:\\

\#\# Seed Questions:\\
{\red [SEED QUESTION 1]}\\
{\red [SEED QUESTION 2]}\\
{\red [SEED QUESTION 3]}\\

\#\# Tabular Data:\\
{\red [TABLE DESCRIPTION]}\\

\#\# Task Requirements:\\
\hspace{15pt}

Please refer to both the tabular data and the seed questions to generate 3 new questions. Ensure that the generated questions meet the following criteria:\\
1.	Each question must be directly related to the provided tabular data and answerable based on that data.\\
2.	The difficulty level of the new questions should be comparable to that of the seed questions in terms of reasoning depth.\\
3.	Each question must be semantically complete, clearly worded, and consistent with natural human expression.\\
4.	The questions should not be exact repetitions or too similar to the seed questions—ensure novelty and diversity in phrasing.\\
5.	The three questions must not be similar to each other; the questioning strategies should be flexible and varied.\\

\#\# Output Format:\\

\hspace{15pt}
\begin{minipage}{\dimexpr\linewidth-15pt}
Question 1:...\\
Question 2:...\\
Question 3:...\\
\end{minipage}

Do not include any other content in the output.

\end{tcolorbox}

\begin{tcolorbox}[sidebyside, sidebyside align=top seam, width=\linewidth, colback=gray!20, colframe=white, colbacktitle=white, coltitle=white, breakable, arc=0mm, left=0mm, right=0mm]
\small

You are a skilled question designer with deep experience in generating reasoning-based questions from tabular data. Below are 3 example seed questions along with a description of the table they are based on. Your goal is to craft 3 new questions that reflect the same level of reasoning complexity and use similar logic types. Each question must be fully grounded in the table and answerable without external knowledge.\\

\#\# Seed Questions:\\
{\red [SEED QUESTION 1]}\\
{\red [SEED QUESTION 2]}\\
{\red [SEED QUESTION 3]}\\

\#\# Table Description:\\
{\red [TABLE DESCRIPTION]}\\

\#\# Guidelines:\\
\hspace{15pt}

Use the table description and the seed questions as references. The questions you generate should fulfill the following conditions:\\
1.	Each question must be clearly tied to the information in the table and solvable using the table alone.\\
2.	Maintain a medium reasoning level—such as involving one layer of condition, filtering, grouping, or a basic calculation.\\
3.	The wording of each question should be natural, specific, and unambiguous.\\
4.	Avoid closely mirroring the structure or content of the seed questions; introduce variety in phrasing and focus.\\
5.	The three questions should each highlight a different reasoning pattern or aspect of the data.\\

\#\# Output Format:\\

\hspace{15pt}
\begin{minipage}{\dimexpr\linewidth-15pt}
Question 1:...\\
Question 2:...\\
Question 3:...\\
\end{minipage}

Only output the three questions—no extra explanation or commentary.

\end{tcolorbox}

\begin{tcolorbox}[sidebyside, sidebyside align=top seam, width=\linewidth, colback=gray!20, colframe=white, colbacktitle=white, coltitle=white, breakable, arc=0mm, left=0mm, right=0mm]
\small

You are a tabular data reasoning expert. Given the 3 seed questions and a description of a table, your task is to generate 3 new questions that demonstrate a similar level of analytical thinking. The questions should be based only on the data provided in the table and should encourage intermediate-level reasoning.\\

\#\# Seed Questions:\\
{\red [SEED QUESTION 1]}\\
{\red [SEED QUESTION 2]}\\
{\red [SEED QUESTION 3]}\\

\#\# Table Description:\\
{\red [TABLE DESCRIPTION]}\\

\#\# Generation Criteria:\\
\hspace{15pt}

Create 3 new questions that meet all of the following criteria:\\
1.	All questions must be directly answerable from the data provided in the table.\\
2.	Aim for medium complexity—questions that may involve conditional selection, simple aggregation, or single-layer logic chains.\\
3.	Ensure natural phrasing and semantic completeness for each question.\\
4.	Do not repeat patterns from the seed questions; diversify the questioning angles.\\
5.	Each question should target a different aspect or relationship in the table to ensure coverage diversity.\\

\#\# Output Format:\\

\hspace{15pt}
\begin{minipage}{\dimexpr\linewidth-15pt}
Question 1:...\\
Question 2:...\\
Question 3:...\\
\end{minipage}

Please return only the questions—exclude any analysis or notes.

\end{tcolorbox}

\noindent The 5 Seed Questions are shown below:

\begin{tcolorbox}[sidebyside, sidebyside align=top seam, width=\linewidth, colback=gray!20, colframe=white, colbacktitle=white, coltitle=white, breakable, arc=0mm, left=0mm, right=0mm]
\small

Seed Question 1: What is the total estimated duration of all ongoing projects? \\

Seed Question 2: What is the sum of earnings per share and net assets per share for Gree Electric? \\

Seed Question 3: How many projects have outstanding payments exceeding 10,000 yuan? \\

Seed Question 4:  What is the ratio of the current price-to-earnings (P/E) ratio to the price-to-book (P/B) ratio for Kweichow Moutai?\\

Seed Question 5:  In the 2017 early first batch undergraduate science admissions in Guangdong Province, which university had the highest minimum score among first-choice applicants?
\end{tcolorbox}

\subsubsection{Hard Question Prompt}

The three prompt templates in the prompt library for hard question generation are shown below:
\begin{tcolorbox}[sidebyside, sidebyside align=top seam, width=\linewidth, colback=gray!20, colframe=white, colbacktitle=white, coltitle=white, breakable, arc=0mm, left=0mm, right=0mm]
\small

You are a professional expert in question generation based on tabular data, skilled in uncovering latent value from complex datasets and designing logically coherent, analytically rich question tasks. You will be provided with 3 seed questions and the corresponding table data. Please carefully read the following content:\\

\#\# Seed Questions:\\
{\red [SEED QUESTION 1]}\\
{\red [SEED QUESTION 2]}\\
{\red [SEED QUESTION 3]}\\

\#\# Tabular Data:\\
{\red [TABLE DESCRIPTION]}\\

\#\# Task Requirements:\\
\hspace{15pt}

With reference to both the tabular data and the style of the seed questions, please generate 3 new questions. Ensure that each question meets the following criteria:\\
1.	Mathematical Reasoning: Aim to design questions that involve computations such as summation, averaging, difference, and ratio analysis.\\
2.	Sorting and Ranking: Include questions that require sorting or ranking by one or more dimensions, either ascending or descending.\\
3.	Horizontal and Vertical Reasoning: Create questions that focus not only on a single row or column, but also involve reasoning across multiple rows and columns.\\
4.	Data Mining Perspective: Design questions from angles such as extreme value comparison or proportion analysis, rather than simple data lookup.\\
5.	Logical Depth: Questions should reflect inferential reasoning, such as associating multiple variables or analyzing changes over time.\\
6. STEM-Oriented Complexity: The new question should be a science/math-type problem involving multi-step reasoning and calculation.\\
7.	Explicit Subject: Each question must have a clear subject, which should correspond to an element present in the tabular data.\\
8.	Computational Feasibility: Ensure that each question can be answered through Python-based tabular data processing. Avoid theoretical questions that cannot be answered through computation.\\

\#\# Output Format:\\

\hspace{15pt}
\begin{minipage}{\dimexpr\linewidth-15pt}
Question 1:...\\
Question 2:...\\
Question 3:...\\
\end{minipage}

Do not include any other content in the output.
\end{tcolorbox}

\begin{tcolorbox}[sidebyside, sidebyside align=top seam, width=\linewidth, colback=gray!20, colframe=white, colbacktitle=white, coltitle=white, breakable, arc=0mm, left=0mm, right=0mm]
\small
You are a high-level expert in generating advanced analytical questions based on structured tabular data. You will be provided with 3 seed questions and a table description. Your task is to generate 3 new, computationally feasible questions that reflect similar reasoning complexity while offering novel perspectives.\\

\#\# Seed Questions:\\
{\red [SEED QUESTION 1]}\\
{\red [SEED QUESTION 2]}\\
{\red [SEED QUESTION 3]}\\

\#\# Table Description:\\
{\red [TABLE DESCRIPTION]}\\

\#\# Question Design Criteria:\\
\hspace{15pt}

Each question you create must fulfill the following constraints:\\
1.	Involve multi-step analytical operations such as multi-attribute comparisons, derived metric calculations, or multi-column aggregations.\\
2.	Encourage reasoning across multiple dimensions—horizontal (row-based), vertical (column-based), or temporal (if applicable).\\
3.	Introduce complexity via ranking, conditional filtering, or chaining multiple logical steps.\\
4.	Explore relationships, changes, or patterns (e.g., growth rates, top/bottom contributors, proportional shifts).\\
5.	Ensure clarity in wording with a concrete subject tied to table attributes.\\
6.	Questions must be answerable purely via tabular operations (e.g., using pandas or SQL). No open-ended or subjective items.\\

\#\# Output Format:\\

\hspace{15pt}
\begin{minipage}{\dimexpr\linewidth-15pt}
Question 1:...\\
Question 2:...\\
Question 3:...\\
\end{minipage}

Exclude any explanatory notes or extra content.

\end{tcolorbox}

\begin{tcolorbox}[sidebyside, sidebyside align=top seam, width=\linewidth, colback=gray!20, colframe=white, colbacktitle=white, coltitle=white, breakable, arc=0mm, left=0mm, right=0mm]
\small

You are a specialist in high-difficulty question generation from structured tables, with a focus on constructing multi-step, computation-driven analytical tasks. Given 3 example questions and a table description, generate 3 new, non-trivial questions that challenge logical reasoning and data synthesis.\\

\#\# Seed Questions:\\
{\red [SEED QUESTION 1]}\\
{\red [SEED QUESTION 2]}\\
{\red [SEED QUESTION 3]}\\

\#\# Tabular Context:\\
{\red [TABLE DESCRIPTION]}\\

\#\# Instructions:\\
\hspace{15pt}

Please ensure each generated question satisfies the following conditions:\\
1.	Requires combining information across multiple rows/columns, potentially involving filtering, aggregation, ranking, or cross-variable comparison.\\
2.	Includes math-based reasoning such as growth computation, ratio inference, delta tracking, or trend evaluation.\\
3.	Reflects real-world data analysis challenges, such as identifying maximum-impact items or tracing category-wise contributions.\\
4.	The questions must contain a well-defined analytical focus and be executable in practice via common data processing tools.\\
5.	Each question should be distinct in logic path from both the seed questions and the other generated questions.\\

\#\# Output Format:\\

\hspace{15pt}
\begin{minipage}{\dimexpr\linewidth-15pt}
Question 1:...\\
Question 2:...\\
Question 3:...\\
\end{minipage}

Return only the 3 questions. Do not add any explanation or metadata.

\end{tcolorbox}

\noindent The 5 Seed Questions are shown below:

\begin{tcolorbox}[sidebyside, sidebyside align=top seam, width=\linewidth, colback=gray!20, colframe=white, colbacktitle=white, coltitle=white, breakable, arc=0mm, left=0mm, right=0mm]
\small

Seed Question 1: In the 2015 pre-interview qualification review list for public institution recruitment in Dandong City, calculate the standard deviation of test scores for each position and identify the position with the greatest score fluctuation. \\

Seed Question 2: In 2023, first filter out cities with fewer than 2 million insured individuals, then identify the city with the highest share of cumulative balance relative to the provincial total. \\

Seed Question 3: Based on the table data, calculate the average difference between the designed and measured longitudinal elevation values for each stake location. Then rank the locations by this average difference in descending order and identify the location with the largest average discrepancy. \\

Seed Question 4: Sort the monthly outbound quantities in ascending order, then calculate the difference between the average outbound quantity of the last three months and that of the first three months. \\

Seed Question 5: Calculate the compound annual growth rate (CAGR) of the total import and export value of agricultural products in Shandong Province from 2010 to 2023, and compare it with the CAGR of general trade import and export value. Specifically, compute the total import and export values for both categories by year, then calculate and compare their CAGRs to determine which category grew faster.\\

\end{tcolorbox}

\subsection{Details of Procedure for Question Annotation}
\label{Appd:question_annotation_procedure}

We randomly assign each question to two annotators, whose selection criteria and qualifications are detailed in Section~\ref{Appd:annotation}.

Each annotator assesses the quality of question candidate based on the following aspects: \textbf{a) scope compliance}: the question must be answerable using tabular data, without requiring any extraneous domain knowledge. Temporal and spatial references must be strictly confined within the boundaries of the dataset. \textbf{b) thematic focus}: the question should concentrate on a single analytical dimension to derive evidence-bound conclusions, rather than enabling the generation of multi-thematic reports across divergent analytical directions. \textbf{c) conceptual distinctiveness}: multiple questions derived from the same table must address non-overlapping thematic aspects with clearly differentiated analytical objectives. \textbf{d) structural and cognitive diversity}: across a set of questions for the same table, annotators must vary the linguistic phrasing and the logical operations required (e.g., comparison, aggregation, trend analysis, or extremum identification). Avoid repetitive sentence templates; each question should represent a distinct way of querying the data to ensure the model's robustness across diverse natural language inputs.

In cases where the evaluation results of the two annotators are inconsistent, the results will be handed over to a third annotator for the final judgment. Through this rigorous quality assurance procedure, we obtained 5,523 high-quality, comprehensive questions.

\subsection{Prompt for Answer Generation}
\label{Appd:answer_generation}

\begin{tcolorbox}[sidebyside, sidebyside align=top seam, width=\linewidth, colback=gray!20, colframe=white, colbacktitle=white, coltitle=white, breakable, arc=0mm, left=0mm, right=0mm]
\small

You are a data analyst proficient in Python. Your task is to write executable Python code to parse tables and then answer questions.\\

\#\# Requirements:\\
\hspace{15pt}

1. Based on the question, write out your analysis approach and methodology, and then write Python code according to this method.\\
2. Please strictly follow the given file path and table description to generate the code.\\
3. Generate only one code block, starting strictly with ```python and ending with ```.\\
4. Your analysis must be completely based on the table data. If the user's question is unrelated to data analysis, please politely decline.\\
5. You need to generate executable code. If there are results to display, please store them in the answer function and display them using print.\\
6. Ensure to use the \texttt{pd.read\_csv} function from the Python library to read the table file path provided for data processing. If multiple tables need to be read, pay attention to using the file path for variable name differentiation.\\
7. During the code generation process, do not convert the data into DataFrame format. Be sure to use the \texttt{pd.read\_csv} function to read the table content.\\

The following is the table information provided:\\
{\red [TABLE DESCRIPTION]}\\

Ensure that the final answer is the last line of the Python code, and can only appear in the form of print(f'{{answer}}'), with no other forms.\\

Let's think step by step, and then generate Python code to analyze the table and display the final answer to the question.\\\

\# \# Input question:\\
{\red [QUESTION]}\\
\end{tcolorbox}

\subsection{Details of Procedure for Answer Annotation}
\label{Appd:answer_annotation_procedure}
We assign each <table, question> pair to two annotators, whose selection criteria and qualifications are detailed in Section A.3. The two annotators are assigned answer annotation tasks with priority given to their respective domain expertise to ensure specialized annotators handle data within their professional fields to guarantee annotation reliability.

Each annotator need to strictly adhere to the following criteria:
a) Scope Compliance: the question must be answerable solely using tabular data without external domain knowledge. Temporal and spatial references must be strictly confined to the dataset boundaries.
b) Clarity \& Conciseness: the annotated answer should be straightforward and concise, avoiding redundant content or thematic digression.
c) Linguistic Fluency: the annotated answer must maintain grammatical coherence and fluency, free from obvious language errors.

In cases where the evaluation results of the two annotators are inconsistent, the results will be handed over to a third annotator for the final judgment. By this rigorous quality assurance procedure,
we obtain 5,523 high-quality <table, question, answer> triples.

\subsection{Details of Procedure for Reasoning Process Annotation}
\label{Appd:question_reasoning_annotaion_procedure}
We assign each <table, question, reasoning process> triples to two annotators, with filtering and selecting the most representative high quality reasoning process. The two annotators are assigned the whole reasoning process annotation tasks with priority given to their respective domain expertise to ensure specialized annotators handle data within their professional fields to guarantee annotation reliability. 

To guarantee the correctness of the reasoning traces produced by the LLM, annotations must adhere strictly to the following requirements: 1) Table-Answerability: the reasoning process must be answerable solely from the table data; no external domain knowledge is permitted. Temporal or spatial references must be strictly confined to the dataset. 2) Data Accuracy: all data citations must be exact matches to the original table content, with numerical values accurate within a ±0.5 \% floating-point tolerance. Post-computation values must be validated for formulaic correctness. Cross-table references must be verified for both source table and field accuracy. 3) Logical Completeness: whether produced under thinking-mode or no-thinking-mode, the reasoning process must form a complete logical chain. carefully verify: a) header-field correctness,  b) numerical-computation logic, c) especially for multi-table reasoning, the correct mapping of join fields, d) irrelevant or erroneous content must be removed. 4) Codes and Comments: all generated code must pass syntax checks and must be re-examined for executability (even if previously validated) and for the accuracy of comments, ensuring full alignment between comments and code logic. 5) Semantic Consistency: every reasoning process must pass a two-stage verification: a) does it precisely address the core requirement of the original question?  b) does it introduce any derivative conclusions not explicitly requested by the question? Such extraneous conclusions must be deleted.

In cases where the evaluation results of the two annotators are inconsistent, the results will be handed over to a third annotator for the final judgment. By this rigorous quality assurance procedure, we obtain two annotated SFT datasets containing 1932 and 1932 <table, question, reasoning process> triples, respectively.

\subsection{Domain and Sub-domain of ReasonTabQA}
\label{Appd:domain_distribution}
The 7 domains and 30 sub-domains in ReasonTabQA are shown in Table~\ref{Tab:domains}.

\begin{table}[ht]
   \centering
    \resizebox{\linewidth}{!}{
    \begin{tabular}{ll}
    \toprule
    \textbf{Domains} & \textbf{Sub-domains} \\ 
    \midrule
    \makecell[l]{Sales and Marketing} & \makecell[l]{Tourism and Hospitality Services;\\ Food and Beverage Services;\\ Digital Marketing and Social Media;\\ Market Research and Consumer Behavior}  \\ 
    \midrule
    \makecell[l]{Manufacturing and \\Automotive industry} & \makecell[l]{Electronics and Automation Manufacturing;\\ Chemical Engineering and Advanced Materials;\\ Energy Production and Power Systems;\\ Automotive Manufacturing and Mobility Solutions;\\ Industrial Machinery and Heavy Equipment}  \\ 
    \midrule
    \makecell[l]{BI and ERP} & \makecell[l]{Business Management;\\Retail Trade and E-commerce Platforms;\\ Enterprise Resource Planning Systems;\\ Customer Relationship Management}  \\ 
    \midrule
    \makecell[l]{Supply chain} & \makecell[l]{Telecommunications and IT Infrastructure;\\ Transportation Networks and Logistics Management;\\ Procurement and Supplier Relations;\\ Global Trade and Customs Compliance}  \\ 
    \midrule
    \makecell[l]{Healthcare and\\Environmental protection} & \makecell[l]{Healthcare Systems and Public Health; \\Environmental Protection;\\ Agricultural Production and Forestry Management; \\Marine Resources and Fisheries Management}  \\ 
    \midrule
    \makecell[l]{Science and Education} & \makecell[l]{Education and Scientific Research; \\STEM Education and Curriculum Development; \\Academic Research Infrastructure and Laboratory Management;\\ E-Learning and Educational Technology;\\ Language Training and Cultural Exchange}  \\ 
    \midrule
    Finance and Banking & \makecell[l]{Economic Development and International Trade; \\Banking and Financial Services;\\ Investment and Wealth Management;\\ Fintech and Blockchain Technologies}  \\ 
    \bottomrule
    \end{tabular}}
    \captionof{table}{The 7 domains and 30 sub-domains in ReasonTabQA}
    \label{Tab:domains}
\end{table}

\section{Implementation Details for Experiments}

\subsection{Unified Prompt Template for Table Reasoning}
\label{Sec:reasoning_prompt}
\begin{tcolorbox}[sidebyside, sidebyside align=top seam, width=\linewidth, colback=gray!20, colframe=white, colbacktitle=white, coltitle=white, breakable, arc=0mm, left=0mm, right=0mm]
\small

You are a data analyst proficient in Python. Your task is to write executable Python code to parse tables and then answer questions.\\

\#\# Requirements:\\
\hspace{15pt}

1. Based on the question, write out your analysis approach and methodology, and then write Python code according to this method.\\
2. Please strictly follow the given file path and table description to generate the code.\\
3. Generate only one code block, starting strictly with ```python and ending with ```.\\
4. Your analysis must be completely based on the table data. If the user's question is unrelated to data analysis, please politely decline.\\
5. You need to generate executable code. If there are results to display, please store them in the answer function and display them using print.\\
6. Ensure to use the \texttt{pd.read\_csv} function from the Python library to read the table file path provided for data processing. If multiple tables need to be read, pay attention to using the file path for variable name differentiation.\\
7. During the code generation process, do not convert the data into DataFrame format. Be sure to use the \texttt{pd.read\_csv} function to read the table content.\\

The following is the table information provided:\\
{\red [TABLE DESCRIPTION]}\\

Ensure that the final answer is the last line of the Python code, and can only appear in the form of print(f'{{answer}}'), with no other forms.\\

Let's think step by step, and then generate Python code to analyze the table and display the final answer to the question.\\\

\# \# Input question:\\
{\red [QUESTION]}\\
\end{tcolorbox}



\begin{figure*}[ht]
    \centering 
    \includegraphics[width=0.9\textwidth]{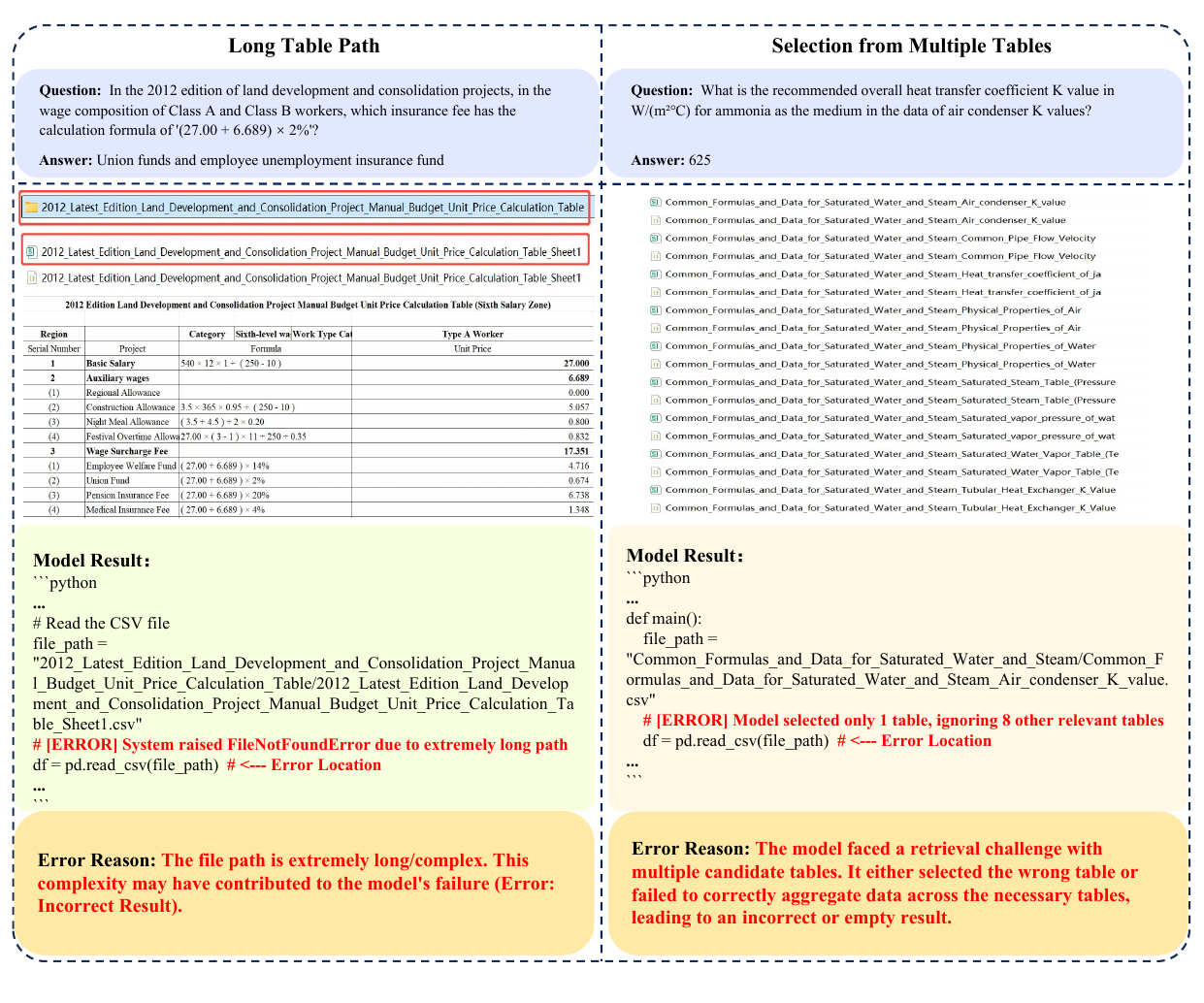} 
    \caption{An example illustrating erroneous file path generation even by the best method Gemini-3-Pro-Preview in scenarios involving long table path or multiple tables, leading to execution failure}
    \label{Fig:case_study_filepath}
\end{figure*}

\begin{figure*}[ht]
    \centering 
    \includegraphics[width=0.9\textwidth]{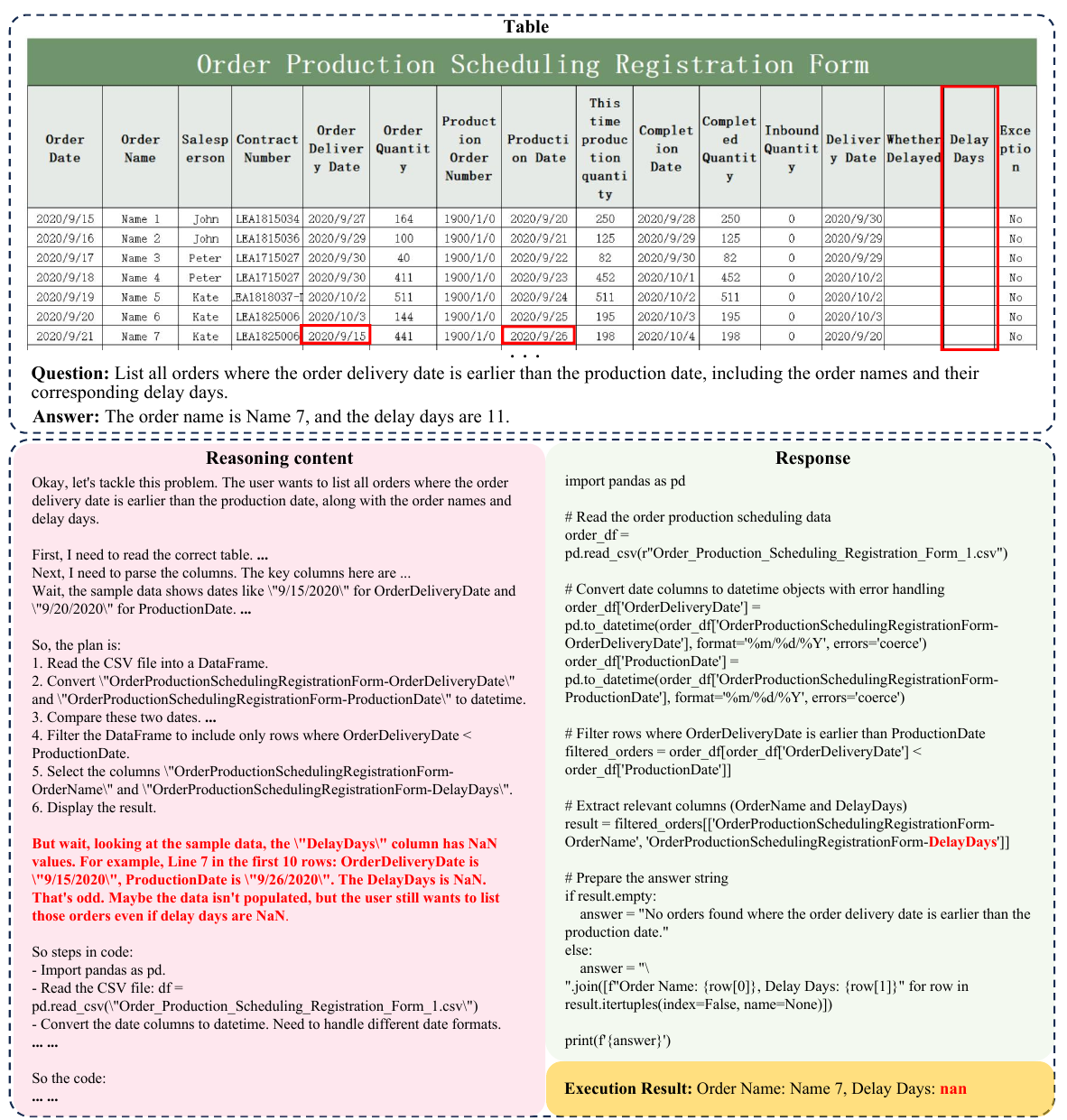} 
    \caption{An example illustrating an original table and its corresponding report generated even by the best method Gemini-3-Pro-Preview, with critical error highlighting.}
    \label{Fig:case_study_error}
\end{figure*}

\begin{figure*}[h]
    \centering 
    \includegraphics[width=0.9\textwidth]{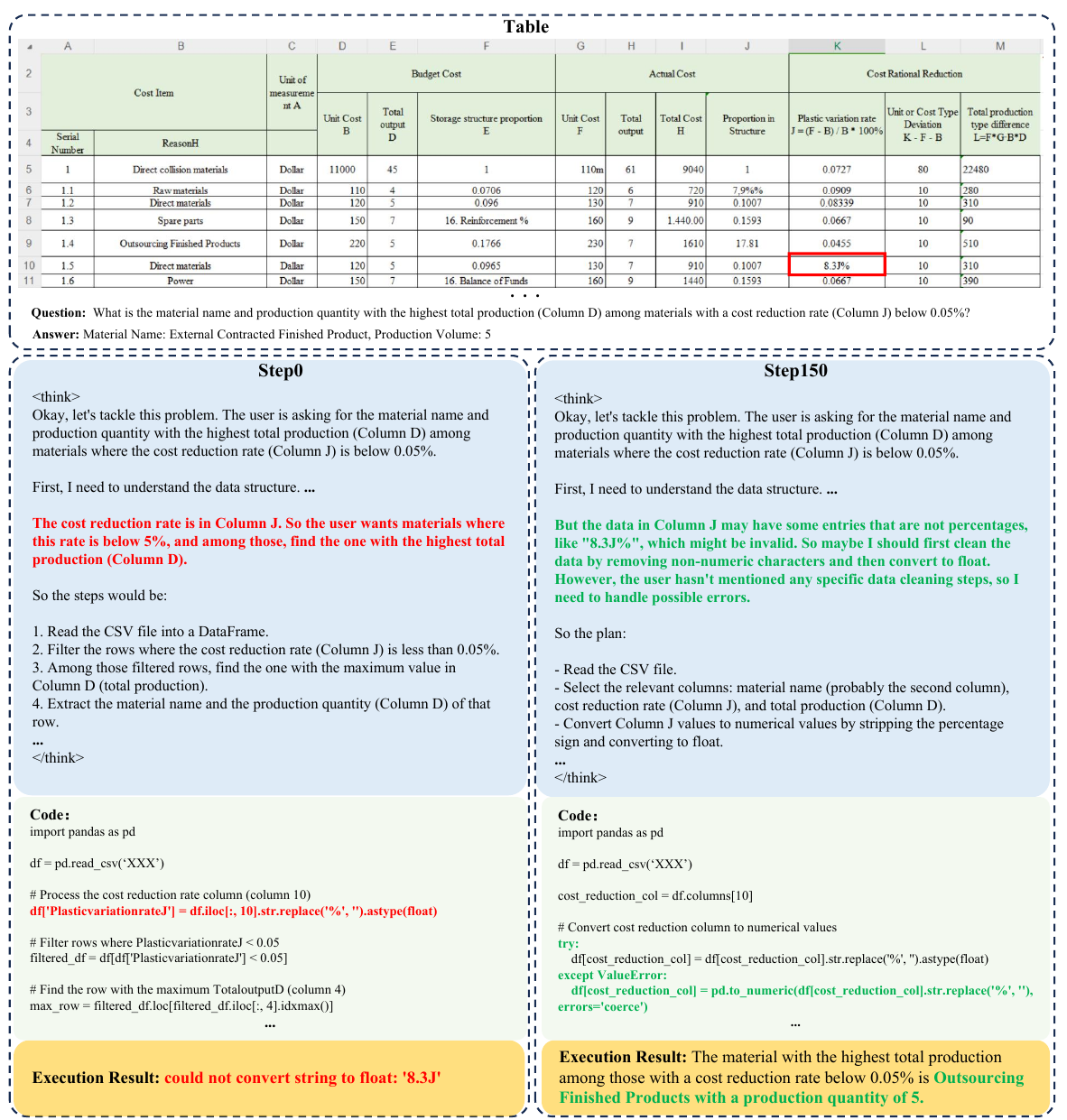} 
    \caption{Comparison of reasoning process and code execution results before and after tabCodeRL on the same case}
    \label{Fig:case_study_rl}
\end{figure*}

\subsection{Analysis of Detailed Case Study} 
\label{Appd:case_study}

Our analysis of the top-performing model (Gemini-3-Pro-Preview) reveals a critical limitation in its reasoning capability and code generation. 
When handling long file names or scenarios involving multiple tables, the model occasionally produces code with incorrect or incomplete file paths, leading to execution failures, as shown in Figure~\ref{Fig:case_study_filepath}. According to statistical analysis, file path-related issues account for approximately 24.29\% of the identified bad cases.

In another case shown in Figure \ref{Fig:case_study_error}, although the model’s reasoning process correctly identified missing values in the "Delay Days" column, it failed to utilize relevant available columns (e.g., Order Delivery Date and Delivery Date) to computationally derive the Delay Days metric. This logical gap resulted in null (NaN) outputs during code execution, despite the presence of sufficient data to support the inference.

As shown in the Figure~\ref{Fig:case_study_rl}, the model trained with TabCodeRL generates more robust and higher-quality code. Specifically, the enhanced model demonstrates improved capability in detecting ill-formatted or anomalous cells (e.g., "8.3J\%" in numerical column) within tabular data and employs defensive programming techniques (e.g., try-except blocks, errors='coerce' in pandas operations) to prevent runtime failures. This results in more resilient code execution, particularly when handling noisy or inconsistent real-world industrial datasets.

\subsection{URLs of Closed-source Models}
The URLs of the closed-source models are as shown in the Table~\ref{Tab:api}.
\begin{table}[ht!]
   \centering
    \resizebox{0.8\linewidth}{!}{
    \begin{tabular}{ll}
    \toprule
    \textbf{Model} & \textbf{URL} \\ 
    \midrule
    \makecell[l]{Claude-4.0-Sonnet} & \makecell[l]{https://www.anthropic.com}  \\ 
    \midrule
    \makecell[l]{Claude-Opus-4.5} & \makecell[l]{https://www.anthropic.com}  \\ 
    \midrule
    \makecell[l]{Gemini3-Pro-Preview} & \makecell[l]{https://deepmind.google}  \\ 
    \midrule
    \makecell[l]{Doubao-1.5thinking-Pro} & \makecell[l]{https://www.volcengine.com} \\ 
    \midrule
    \makecell[l]{GPT-4o} & \makecell[l]{https://openai.com}  \\ 
    \midrule
    \makecell[l]{OepnAI o1-mini} & \makecell[l]{https://openai.com}  \\ 
    \midrule
    \makecell[l]{GPT5.2} & \makecell[l]{https://openai.com}  \\ 
    \bottomrule
    \end{tabular}}
\captionof{table}{The URLs of closed-source models used in experiment}
\label{Tab:api}
\end{table}

\section{Details for payment and GPU hours}
We pay each annotator a daily remuneration of \$40. We paid a total of \$2500 for calling various LLMs API interfaces. We use 16 A100 40G GPUs for inference, which took a total of 25 hours.

\end{document}